\documentclass[journal, twoside]{IEEEtran}
\usepackage{times}
\usepackage[utf8]{inputenc}

\usepackage{amsmath,amssymb,amsfonts}
\let\labelindent\relax
\usepackage{enumitem}
\usepackage{stmaryrd}
\usepackage{multicol}
\usepackage[pdftex]{graphicx}
\usepackage[binary-units=true]{siunitx}
\usepackage[colorlinks,bookmarksopen,bookmarksnumbered,citecolor=black,urlcolor=black]{hyperref}
\hypersetup
{
	pdftitle = {A Feasibility-Driven Approach to Control-Limited DDP},
	pdfauthor = {Carlos Mastalli},
    pdfkeywords = {feasibility, optimal control, differential dynamic programming, control limits, direct transcription, robotics},
	pdftoolbar = true,
	colorlinks = true,
	linkcolor = black,
	citecolor = black,
	urlcolor = black,
}
\usepackage[linesnumbered, ruled, vlined]{algorithm2e}
\usepackage{mwe}
\usepackage{booktabs}
\usepackage{pifont}
\usepackage{glossaries}
\glsdisablehyper
\newacronym{hjb}{HJB}{Hamilton-Jacobi-Bellman}
\newacronym{pmp}{PMP}{Pontryagin's maximum principle}
\newacronym{fonc}{FONC}{first-order necessary condition}
\newacronym{ocnc}{OCNC}{optimal control necessary conditions}
\newacronym{oc}{OC}{optimal control}
\newacronym{nmpc}{nMPC}{nonlinear model predictive control}
\newacronym{mcoc}{MCOC}{multi-contact optimal control}
\newacronym{ilqr}{iLQR}{iterative linear-quadratic regulator}
\newacronym{lqr}{LQR}{linear-quadratic regulator}
\newacronym{gn}{GN}{Gauss-Newton}
\newacronym{qp}{QP}{quadratic programming}
\newacronym{lq}{LQ}{linear quadratic}
\newacronym{slq}{SLQ}{sequential linear quadratic}
\newacronym{sqp}{SQP}{sequential quadratic programming}
\newacronym{nlp}{NLP}{nonlinear programming}
\newacronym{ddp}{DDP}{differential dynamic programming}
\newacronym{fddp}{FDDP}{feasibility-driven differential dynamic programming}
\newacronym{com}{CoM}{center of mass}
\newacronym{crocoddyl}{Crocoddyl}{Contact RObot COntrol by Differential DYnamic Library}
\newacronym{kkt}{KKT}{Karush-Kuhn-Tucker}
\newacronym{rnea}{RNEA}{recursive Newton-Euler algorithm}
\newacronym{aba}{ABA}{articulated body algorithm}
\newacronym{lf}{LF}{left front}
\newcommand{\orcid}[1]{\href{https://orcid.org/#1}{\includegraphics[width=0.6em]{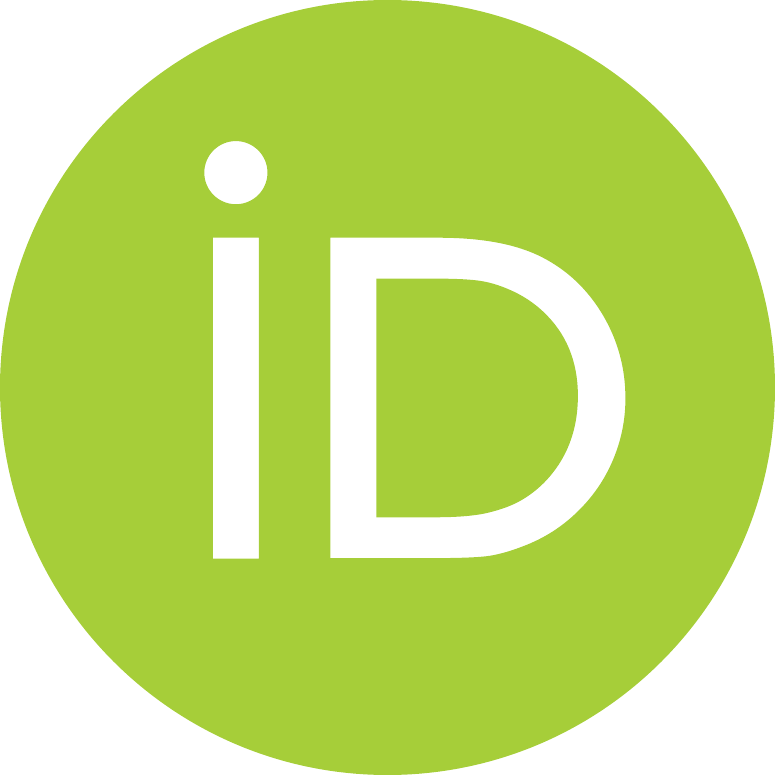}}}
\newcommand{\sref}[1]{Section~\ref{#1}}
\newcommand{\fref}[1]{Fig.~\ref{#1}}
\newcommand{\eref}[1]{Eq.~(\ref{#1})}
\newcommand{\tref}[1]{Table~\ref{#1}}
\newcommand{\aref}[1]{Algorithm~\ref{#1}}
\newlength{\tempdima}
\newcommand{\rowname}[1]{\rotatebox{0}{\makebox[\tempdima][c]{(#1)}}}

\newcommand{\xmark}{\ding{55}}

\begin{document}

\title{A Feasibility-Driven Approach\\to Control-Limited DDP}
\author{Carlos Mastalli\,\orcid{0000-0002-0725-4279}\quad Wolfgang Merkt\,\orcid{0000-0003-3235-4906}\quad Josep Marti-Saumell\,\orcid{0000-0001-7981-7942}\\
\quad\quad Henrique Ferrolho\,\orcid{0000-0003-4307-0028}\quad Joan Sol\`a\,\orcid{0000-0002-2933-3381}\,\quad Nicolas Mansard\,\orcid{0000-0002-8090-0601}\quad Sethu Vijayakumar\,\orcid{0000-0003-0649-7241}
\thanks{%
This research was supported by (1) the European Commission under the Horizon 2020 project Memory of Motion (MEMMO, project ID: 780684), (2) the Engineering and Physical Sciences Research Council (EPSRC) UK RAI Hub for Offshore Robotics for Certification of Assets (ORCA, grant reference EP/R026173/1), and (3) the Alan Turing Institute.
\textit{(Corresponding author: Carlos Mastalli).}
}
\thanks{
Carlos Mastalli is with the Institute of Sensors, Signals and Systems, School of Engineering and
Physical Sciences, Heriot-Watt University, U.K. (e-mail: \href{mailto:c.mastalli@hw.ac.uk}{c.mastalli@hw.ac.uk}).}
\thanks{
Henrique Ferrolho and Sethu Vijayakumar are with the School of Informatics, University of Edinburgh, U.K. (e-mail: \href{mailto:henrique.ferrolho@ed.ac.uk}{henrique.ferrolho@ed.ac.uk}; \href{mailto:sethu.vijayakumar@ed.ac.uk}{sethu.vijayakumar@ed.ac.uk}).}
\thanks{
Wolfgang Merkt is with the Oxford Robotics Institute, University of Oxford, U.K.
(e-mail: \href{mailto:wolfgang@robots.ox.ac.uk}{wolfgang@robots.ox.ac.uk}).}
\thanks{
Josep Marti-Saumell and Joan Sol\`a are with the Institut de Rob\`otica i Inform\`atica Industrial, Universitat Polit\`ecnica de Catalunya, Spain (e-mail: \href{mailto:jmarti@iri.upc.edu}{jmarti@iri.upc.edu}; \href{mailto:jsola@iri.upc.edu}{jsola@iri.upc.edu}).}
\thanks{
Nicolas Mansard is with LAAS-CNRS, France (e-mail: \href{mailto:nicolas.mansard@laas.fr}{nicolas.mansard@laas.fr}).}
}
\markboth{}{Mastalli \MakeLowercase{\textit{et al.}}: A Feasibility-Driven Approach to Control-Limited DDP}

\maketitle

\begin{abstract}
\Gls{ddp} is a direct single shooting method for trajectory optimization.
Its efficiency derives from the exploitation of temporal structure (inherent to optimal control problems) and explicit roll-out/integration of the system dynamics.
However, it suffers from numerical instability and, when compared to direct multiple shooting methods, it has limited initialization options (allows initialization of controls, but not of states) and lacks proper handling of control constraints.
In this work, we tackle these issues with a feasibility-driven approach that regulates the dynamic feasibility during the numerical optimization and ensures control limits.
Our feasibility search emulates the numerical resolution of a direct multiple shooting problem with only dynamics constraints.
We show that our approach (named \textsc{Box-FDDP}) has better numerical convergence than \textsc{Box-\gls{ddp}$^+$} (a single shooting method), and that its convergence rate and runtime performance are competitive with state-of-the-art direct transcription formulations solved using the interior point and active set algorithms available in \textsc{Knitro}.
We further show that \textsc{Box-FDDP} decreases the dynamic feasibility error monotonically---as in state-of-the-art nonlinear programming algorithms.
We demonstrate the benefits of our approach by generating complex and athletic motions for quadruped and humanoid robots.
Finally, we highlight that \textsc{Box-FDDP} is suitable for model predictive control in legged robots.
\end{abstract}

\begin{IEEEkeywords}
    optimal control, differential dynamic programming, feasibility, direct multiple shooting, control limits
\end{IEEEkeywords}

\IEEEpeerreviewmaketitle

\section{Introduction}\label{sec:introduction}
\begin{figure}[t]
    \centering
    \href{https://youtu.be/bOGBPTh_lsU}{\includegraphics[width=\linewidth]{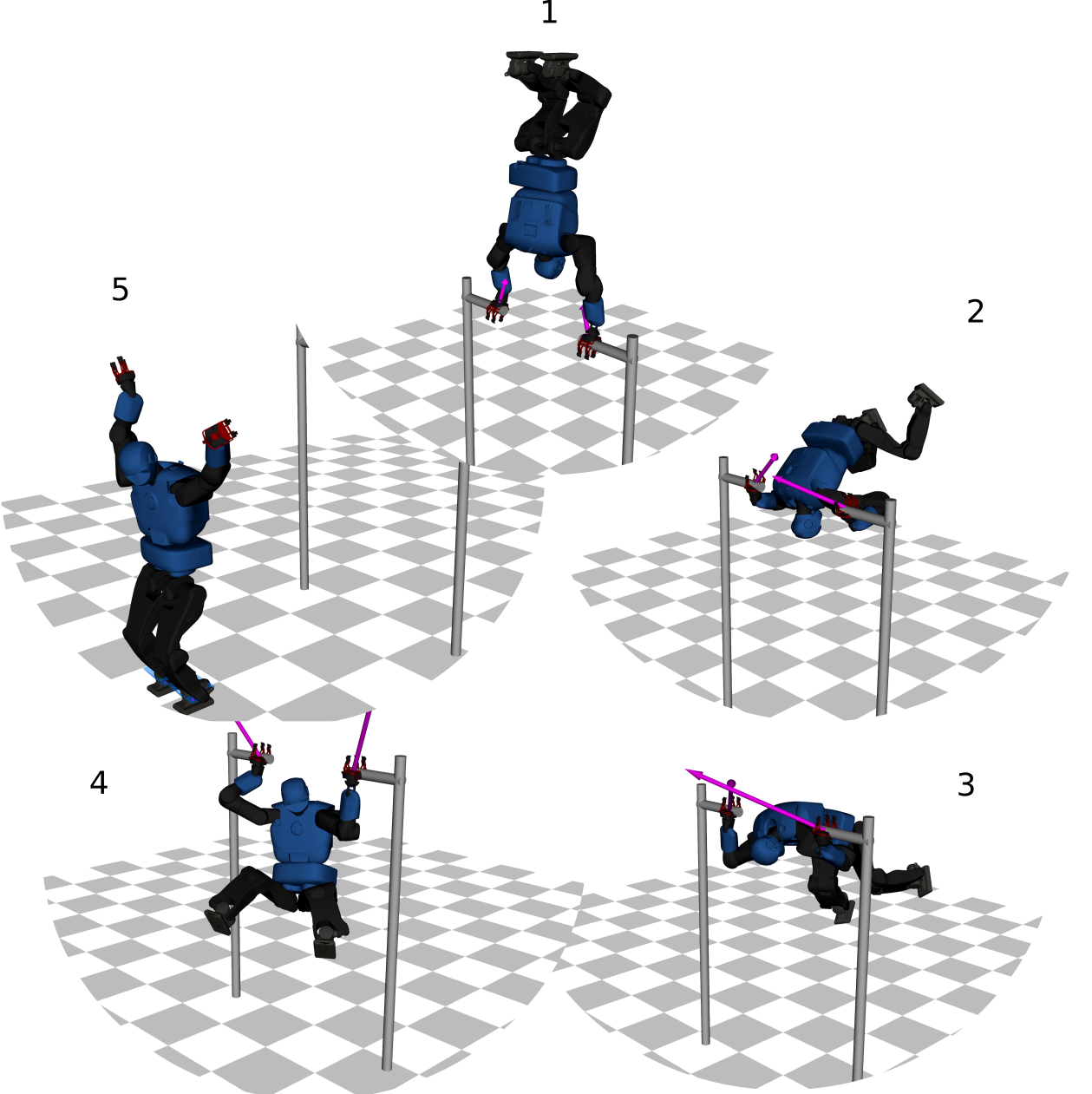}}\vspace{-0.8em}
    \caption{Snapshots of an athletic behavior computed by our feasibility-driven approach with explicit control limits: \textsc{Box-\gls{fddp}}.
    The optimized motion considers the robot's full-body dynamics, joint limits, and friction cone constraints.
    We initialize our algorithm using the default posture and quasi-static torques (as described in the results section).
    We describe three desired actions: grasping the bar at a specific point, raising the feet up-ward a given location (Seq. 1), and landing over a specific placement (Seq. 5).
    The shown behavior emerges without further specification from our problem formulation.
    The arrows describe the magnitude and direction of the contact forces.
    To watch the video, click the figure or see \texttt{\url{https://youtu.be/bOGBPTh_lsU}}.}
    \label{fig:cover}
\end{figure}

\IEEEPARstart{O}{ptimal} control is a powerful tool to synthesize motions and controls through task goals (cost/optimality) and constraints (e.g., system dynamics, interaction constraints).
We can formulate such problems using \textit{direct methods}~\cite{betts-bookoptctrl}, which first discretize over both state and controls, and then optimize using sparse general-purpose \gls{nlp} software such as \textsc{Snopt}~\cite{gill-siam05}, \textsc{Knitro}~\cite{byrd-knitro06}, and \textsc{Ipopt}~\cite{wachter-mp06}.
To ensure that the system state evolves as described by the equations of motion, we define equality constraints in the~\gls{nlp} problem.
However, common algorithms for nonlinear programming can only guarantee the constraint satisfaction at their convergence.
When this occurs, we say that the discretized states are \textit{dynamically} feasible.
Furthermore, despite the advantage of using advanced software for nonlinear programming, these algorithms perform very large matrix factorizations during the computation of the search direction, i.e., the resolution of the~\gls{kkt} problem.
To do so, they use sparse linear solvers such as MA27, MA57, and MA97 (see~\cite{HSL}) that do not exploit the temporal/Markovian structure of optimal control problems efficiently. 
Indeed, the expensive factorizations of these linear solvers limit their practical use to realtime control on reduced models (e.g.,~\cite{wieber-iros02,pardo-ral16,mastalli-icra17,dicarlo-iros18}) or motion planning (e.g.,~\cite{carpentier-icra16,aceituno_cabezas-ral17,winkler-ral18,merkt-iros18,mastalli-tro20}) in robotics.
Furthermore, classical line search methods used in general-purpose \gls{nlp} solvers are less effective than the nonlinear roll-out of the dynamics used in shooting methods as their use increases the number of iterations (cf.~\cite{liao-92}), which is used in recent method for multiple shooting~\cite{giftthaler-iros18}.

\textit{Dynamic programming methods}, which have their foundations in the calculus of variations as indirect methods, have once again attracted attention due to recent results on fast nonlinear model predictive control based on~\gls{ddp} (e.g.,~\cite{tassa-iros12,koenemann-iros15,neunert-ral18,farshidian-humanoids17}).
In particular, there is a significant interest in the \gls{ilqr} algorithm~\cite{li-icinco04} as its \gls{gn} approximation reduces the computation time while having super-linear convergence.
Both~\gls{ilqr} and~\gls{ddp} algorithms perform a Riccati sweep in the backward pass, which incorporates elements that are reminiscent of \gls{pmp}.
For instance, at convergence, the gradient of the value function in the backward pass represents the \textit{costate}; instead, the roll-out of the system dynamics describes the \textit{state integration} step.
This connection was recognized by Bellman's groundbreaking work~\cite{bellman54bull} that established the so-called \gls{hjb} equation in the continuous-time domain.
In contrast to classical direct collocation approaches, these approaches exploit the temporal/Markovian structure of the optimal control problem by solving a sequence of smaller sub-problems derived from Bellman's principle of optimality~\cite{mayne-66}.
This leads to fast and cheap computations due to very small matrix factorizations and effective data cache accesses.
Despite these advantages, both algorithms are unable to handle equality and inequality constraints efficiently.
Furthermore, they have a poor basin of attraction for a good local optimum as it requires a good initialization in order to converge and are prone to numerical instability---commonly recognized challenges for single shooting approaches~\cite{betts-bookoptctrl}.
These undesirable properties are mainly due to the fact that the \gls{ilqr}/\gls{ddp} algorithms implicitly enforce the dynamic feasibility through the system roll-out.

\subsection{Related work}
Trade-offs between feasibility and optimality appear in most of the state-of-the-art nonlinear programming software.
For instance, \textsc{Ipopt} includes a feasibility restoration phase which aims at reducing the constraint violation~\cite{wachter-mp06}.
In \textsc{Knitro}, the progress on both feasibility and optimality is achieved by adding an $\ell^1$-norm penalty term for the constraints in the \textit{merit} function~\cite{byrd-knitro06}.
In fact, by changing the merit function or the line search procedure, we can put emphasis on obtaining feasible solutions \textit{before} trying to optimize them.
Instead, the \gls{ilqr}/\gls{ddp} algorithms do not make this trade-off, as the backward and forward passes do not accept infeasible iterations.
However, recent work on \textit{multiple shooting \gls{ddp}} ~\cite{giftthaler-iros18,mastalli-icra20} has provided ways of handling dynamically infeasible iterations, which we elaborate below.

The multiple shooting variants in~\cite{giftthaler-iros18,mastalli-icra20} are rooted in dynamic programming.
For instance, Giftthaler et al.~\cite{giftthaler-iros18} introduced a \textit{lifted}\footnote{This name is coined by~\cite{albersmeyer-siam10}, and we refer to \textit{gaps} or \textit{defects} produced between multiple shooting nodes.} version of the algebraic Riccati equation that allows initialization of both state and control trajectories; it further accounts for the relinearization required by the dynamics gaps in the backward pass and uses a merit function to balance feasibility and optimality.
In turn, in our previous work~\cite{mastalli-icra20}, we proposed a modification of the forward pass that numerically matches the gap contraction expected by a \textit{direct multiple shooting} method subject to equality constraints only.
It factorizes the~\gls{kkt} matrix via a Riccati recursion and defines the behavior of the defect constraints based on the \gls{fonc} of optimality.\footnote{For more details about the \gls{fonc} of optimality see~\cite{nocedal-optbook}.}
These approaches improve numerical robustness against poor initialization, as they are able to use an initial guess for the state trajectory.
Unfortunately, none of these methods handle inequality constraints such as control limits, with the exception of a recent work that computes squashed control sequences~\cite{marti-iros20}.

There are two main strategies for incorporating arbitrary constraints: active set and penalization methods (as extensively described in~\cite{nocedal-optbook}).
In the robotics community, one of the first successful attempts to incorporate inequality constraints in \gls{ddp} used an active set approach~\cite{tassa-icra14}, which is based on~\cite{ohno-78} -- a pioneering work in the control community.
Concretely, this approach focused on handling control limits during the computation of the backward pass, i.e., in the minimization of the action-value function ($Q-$function),\footnote{In the following section we formally describe the action-value function (i.e.,$Q-$function).} which resembles the control Hamiltonian at convergence (see \cite{kirk-optctrlbook}, Section 3.11).
The method is popularly named \textsc{Box-\acrshort{ddp}}, and the authors also showed a better convergence rate when compared with a squashing function approach.
Later, Xie et al.~\cite{xie-icra17} included general inequality constraints into the $Q-$function and the forward pass.
The method sacrifices the computational performance by including a second quadratic program, which is solved in the forward pass.
However, it still remains faster than solving the same problem using direct collocation with \textsc{Snopt} as reported in~\cite{hargraves-jguidance87}.

Generally speaking, active set methods are suitable for small-size problems (such as minimizing the $Q-$function described above) as their accuracy and speed often outperform other methods.
However, the combinatorial complexity of finding the active set is prohibitive in large-scale optimization problems.
This motivates the development of penalty-based methods, despite their numerical difficulties: ill-conditioning and slow convergence.
To overcome these difficulties, Lantoine and Russell~\cite{lantoin-aiaa08} proposed a method that incorporates an augmented Lagrangian term.
This method was studied in the context of robust thrust optimization, in which the dynamical system has fewer degrees of freedom compared to complex legged robots.
Later, Howell et al.~\cite{howell-19} extended the augmented Lagrangian approach to handle arbitrary inequality constraints for aerial navigation and manipulation problems.
Additionally, the algorithm incorporates an active set projection for solution polishing and is often faster than direct collocation solved with \textsc{Ipopt} or \textsc{Snopt}.

Our work proposes a feasibility-driven search for nonlinear optimal control problems with control limits.
The main motivation of our approach is to increase the algorithm's basins of attraction, by focusing on feasibility instead of focusing solely on efficiency and optimality.
Apart from the control limits and dynamics, we handle all remaining constraints (e.g., state and friction cone) through quadratic penalization, as described in the results section.

\subsection{Contribution}
The main contribution of this work is the first complete study of the numerical properties, behaviors, and guarantees of feasibility-driven search in differential dynamic programming.
It relies on three technical contributions:
\begin{enumerate}[label=(\roman*)]
    \item an original and efficient optimal control algorithm that directly handles control limits (\textsc{Box-\gls{fddp}}),
    \item extensive comparisons against direct transcription and \textsc{Box-\gls{ddp}$^+$} (a single shooting method),
    \item a tutorial that connects the different branches of theory in optimal control, and
    \item an experimental validation of the dynamic feasibility evolution against interior point and active set algorithms for nonlinear programming.
\end{enumerate}

Our approach builds on top of our previous results on feasibility-driven search~\cite{mastalli-icra20}, for which we hereby propose to define two modes in our algorithm: feasibility-driven and control-bounded.
It considers the dynamic feasibility in the forward pass and explicitly incorporates control limits, which does not require a merit function as in~\cite{giftthaler-iros18}.
Additionally, our approach has outstanding numerical capabilities, which allow us to generate motions that go beyond state-of-the-art methods on optimal control or trajectory optimization in robotics, e.g., the athletic maneuver of a humanoid robot shown in~\fref{fig:cover}.

\section{Direct multiple shooting and differential dynamic programming}\label{sec:mocp}
Before describing our approach, we introduce direct multiple shooting, and explain its numerical advantages when compared to single shooting methods such as~\gls{ddp} (\sref{sec:direct_transcription}).
Then, in \sref{sec:connection} we present a unique tutorial that connects the various branches of theory: \gls{kkt}, \gls{pmp}, and~\gls{hjb}.
Additionally, in \sref{sec:boxddp} we describe the salient aspects of original \textsc{Box-\gls{ddp}} proposed by~\cite{tassa-icra14}, and our variant \textsc{Box-\gls{ddp}$^+$}.
This section contains known material, although we believe it contributes (i) to unveil the underlying problems of differential dynamic programming, and (ii) to understand the theoretical foundations of our feasibility-driven approach.

\subsection{Direct multiple shooting for optimal control}\label{sec:direct_transcription}
Without loss of generality, we consider a direct multiple shooting approach for the nonlinear optimal control problem with control bounds in which each shooting segment defines a single timestep:
\begin{equation}
	\label{eq:multiple_shooting_oc_problem}
	\begin{aligned}
			\min_{\mathbf{x}_s,\mathbf{u}_s}
			&
			& & \ell_N(\mathbf{x}_N) + \sum_{k=0}^{N-1} \ell_k(\mathbf{x}_k,\mathbf{u}_k) \\
			& \hspace{-2em}\textrm{s.t.}
			& &   \hspace{-1em}\mathbf{\bar{f}}_0\,\,\,\,\,\,:=\mathbf{x}_0\ominus\tilde{\mathbf{x}}_0=\mathbf{0},\\
            & & & \hspace{-1em}\mathbf{\bar{f}}_{k+1}:=\mathbf{f}(\mathbf{x}_k,\mathbf{u}_k)\ominus\mathbf{x}_{k+1}=\mathbf{0}, &\hspace*{1em} \scriptstyle{\forall k=\{0,1,\cdots,N-1\}}\\
            & & & \hspace{-1em}\mathbf{\underline{u}} \leq \mathbf{u}_k \leq \mathbf{\bar{u}}, &\hspace*{1em}\scriptstyle{\forall k=\{0,1,\cdots,N-1\}}\\
	\end{aligned}
\end{equation}
where the state $\mathbf{x}\in X$ lies in a differential manifold (with dimension $n_x$); the control $\mathbf{u}\in\mathbb{R}^{n_u}$ defines the input commands; $\mathbf{\underline{u}}$,~$\mathbf{\bar{u}}$ are the lower and upper control bounds; $\tilde{\mathbf{x}}_0$ is the initial state of the system; $\ominus$ describes the \textit{difference} operator of the state manifold (notation inspired by~\cite{frese-thesis} that is needed to optimize over manifolds~\cite{gabay82jota}); $N\in\mathbb{N}$ describes the number of nodes (horizon); $\ell_N$, $\ell_k$ are the terminal and running cost functions; and $\mathbf{\bar{f}}_0$, $\mathbf{\bar{f}}_{k+1}\in T_{\mathbf{x}}X$ are the gap residual functions that impose the dynamic feasibility; and $T_{\mathbf{x}}X$ describes the tangent space of the state manifold at the state $\mathbf{x}$.

\begin{figure}%
    \centering
    \includegraphics[width=1.\columnwidth]{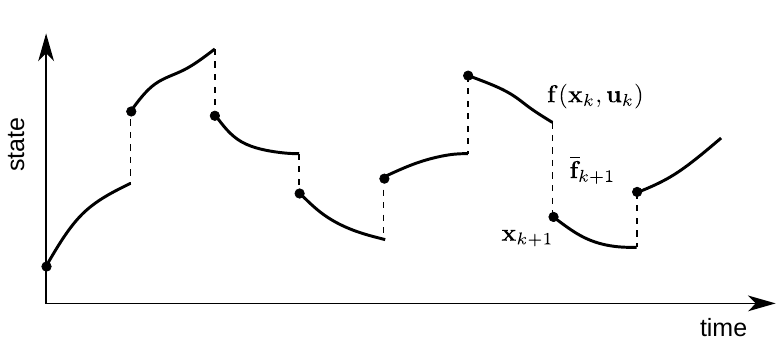}\vspace{-0.8em}
    \caption{A schematic of a direct multiple shooting formulation.
	Different intermediate states are introduced as decision variables.
	A set of equality constraints enforces the dynamics feasibility.
	The black dots represent the initial states used for the system roll-out.
	The dashed lines represent the gaps between the roll-outs.
	Dynamic feasibility is achieved once these gaps are closed.}
    \label{fig:multiple_shooting}
\end{figure}

\eref{eq:multiple_shooting_oc_problem} describes a nonlinear program as the system dynamics are transcribed into a set of algebraic equations with defects in each timestep.
It is possible to extend this notation for cases where shooting segments contain multiple timesteps; however, as seen later, this does not provide any computational benefit, i.e., reduction in the computation time or better distribution of nonlinearities of the dynamics.
\fref{fig:multiple_shooting} depicts the transcription process incorporating state and control trajectories $(\mathbf{x}_s,\mathbf{u}_s)$ as decision variables.
This is in contrast to differential dynamic programming, which only transcribe the control sequence $\mathbf{u}_s$ and obtain $\mathbf{x}_s$ by integrating the system dynamics (i.e., a single shoot).

\subsubsection{Numerical behavior of direct multiple shooting}\label{sec:fddp_kkt}
Algorithms for nonlinear programming aim at finding the \acrfull{kkt} conditions defined by the \gls{fonc} of optimality~\cite{nocedal-optbook}.
This process involves iteratively solving a \gls{kkt} problem (i.e., linear system of equations) until satisfaction of a stopping criterion.
In the \textit{line search} strategy, the solution of this~\gls{kkt} problem provides a search direction $\delta\mathbf{w}_k$, and the selected step length $\alpha$ defines how much the current guess $\mathbf{w}_k$ moves along that direction, i.e., $\mathbf{w}_{k+1} = \mathbf{w}_{k} \oplus \alpha\delta\mathbf{w}_{k}$. 
Note that the \textit{integrator} operator $\oplus$ enables us to optimize over the manifold (as in~\cite{frese-thesis,gabay82jota}), however, it is a feature that general-purpose nonlinear programming libraries often does not have.

We can easily analyze the numerical behavior of direct multiple shooting formulations by focusing on the \gls{kkt} problem for the shooting interval $k$ only.
This is possible because of the temporal/Markovian structure of optimal control problems.
Therefore, when we apply the Newton method on this \gls{kkt} sub-problem together with the Bellman's principle of optimality, we obtain:
\begin{equation}\label{eq:multishooting_kkt}
\begin{bmatrix}
\boldsymbol{\ell}_{\mathbf{xx}_k} & \boldsymbol{\ell}_{\mathbf{ux}_k}^\top & \mathbf{f}^\top_{\mathbf{x}_k} & \\
\boldsymbol{\ell}_{\mathbf{ux}_k} & \boldsymbol{\ell}_{\mathbf{uu}_k} & \mathbf{f}^\top_{\mathbf{u}_k} & \\
\mathbf{f}_{\mathbf{x}_k} & \mathbf{f}_{\mathbf{u}_k} & & -\mathbf{I} \\
& & -\mathbf{I} & \mathcal{V}_{\mathbf{xx}_{k+1}}
\end{bmatrix}
\begin{bmatrix}
\delta\mathbf{x}_k \\ \delta\mathbf{u}_k \\ \boldsymbol{\lambda}^+_{k+1} \\ \delta\mathbf{x}_{k+1}
\end{bmatrix} = -
\begin{bmatrix}
\boldsymbol{\ell}_{\mathbf{x}_k} \\ \boldsymbol{\ell}_{\mathbf{u}_k} \\ \mathbf{\bar{f}}_{k+1} \\ \mathcal{V}_{\mathbf{x}_{k+1}}
\end{bmatrix},
\end{equation}
where \eref{eq:multishooting_kkt} defines the stationary condition (first and second rows) and the primal feasibility (third row) of the \gls{fonc} of optimality, respectively; $\delta\mathbf{x}_k$,~$\delta\mathbf{u}_k$ define the search direction for the primal variables; $\boldsymbol{\lambda}^+_{k+1}$ is the updated Lagrangian multipliers; $\boldsymbol{\ell}_{\mathbf{x}_k}$, $\boldsymbol{\ell}_{\mathbf{u}_k}$, and $\boldsymbol{\ell}_{\mathbf{xx}_k}$, $\boldsymbol{\ell}_{\mathbf{xu}_k}$, $\boldsymbol{\ell}_{\mathbf{uu}_k}$ are the Jacobians and Hessians of the cost function; $\mathbf{f}_{\mathbf{x}_k}$, $\mathbf{f}_{\mathbf{u}_k}$ are the Jacobians of the system dynamics; and $\mathcal{V}_{\mathbf{x}_k}$, $\mathcal{V}_{\mathbf{xx}_k}$ are the gradient and Hessian of the value function.
Note that we apply the \acrfull{gn} approximation as we ignore the Hessian of the system dynamics to avoid expensive tensor-vector multiplications.

When we factorize this system of equations, the resulting search direction always satisfies the dynamics constraints if the Jacobians and Hessians are constant (i.e., a~\acrshort{lqr} problem).
However, if we apply an $\alpha$-step, the gap of the dynamics closes by a factor of $(1-\alpha)$.
We observe this by inspecting the primal feasibility at the next iteration:
\begin{eqnarray}\label{eq:nonlinear_gap_pred}
	\mathbf{\bar{f}}_{k+1}^{i+1} &=& \mathbf{\bar{f}}_{k+1}^{i} - (\delta\mathbf{x}_{k+1} - \mathbf{f_x}_k\delta\mathbf{x}_k - \mathbf{f_u}_k\delta\mathbf{u}_k)\nonumber\\
	&=& (1-\alpha)(\mathbf{f}(\mathbf{x}^i_k,\mathbf{u}^i_k)\ominus\mathbf{x}^i_{k+1}),
\end{eqnarray}
where, by definition in~\eref{eq:multishooting_kkt} (third row), we have that
\begin{equation*}
	\alpha\mathbf{\bar{f}}^i_{k+1} = \delta\mathbf{x}_{k+1}-\mathbf{f_x}_k\delta\mathbf{x}_k-\mathbf{f_u}_k\delta\mathbf{u}_k,
\end{equation*}
with the gap defined as $\mathbf{\bar{f}}^i_{k+1}=\mathbf{f}(\mathbf{x}^i_k,\mathbf{u}^i_k) \ominus \mathbf{x}^i_{k+1}$, and $i$ describes the iteration number.

As described later in~\sref{sec:boxfddp}, injecting this numerical behavior can be interpreted as a feasibility-driven approach for multiple shooting.
However, our approach operates quite differently from classical multiple shooting approaches.
For instance, it does not increase the computation time by defining extra state (decision) variables.
But there is no such thing as a free lunch as our approach cannot \textit{temporarily} increase the defects (e.g., to reduce the cost value) after taking its first full step ($\alpha=1$).

\subsubsection{Advantages of direct multiple shooting}\label{sec:advantages_multshooting}
The rationale for a direct multiple shooting approach (namely, adding $\mathbf{x}_s$ as decision variables) is to distribute the nonlinearities of the dynamics over the entire horizon~\cite{diehl-fmbr06}.
To illustrate this statement, we recognize that integrating over a horizon implies recursively calling \textit{integrator} functions, i.e.,
\begin{eqnarray}
    \mathbf{x}_{k+1} &=& \mathbf{f}(\mathbf{x}_k, \mathbf{u}_k) \nonumber\\
                     &=& \mathbf{f}(\mathbf{f}(\mathbf{x}_{k-1}, \mathbf{u}_{k-1}), \mathbf{u}_k) \nonumber\\
                     &=& \vdots \nonumber\\
                     &=& \mathbf{f}(\mathbf{f}(\cdots \mathbf{f}(\mathbf{x}_0,\mathbf{u}_0), \mathbf{u}_{k-1}), \mathbf{u}_k),
\end{eqnarray}
in which the nonlinearity increases along the horizon.
This means that the local prediction, constructed by the derivatives of the nonlinear system in the~\gls{kkt} problem, becomes more inaccurate as the horizon increases.
This is a well recognized drawback of single shooting approaches, and certainly a numerical limitation of the \textsc{Box-\gls{ddp}$^+$} algorithm described below.

\subsection{Connection between~\gls{kkt},~\gls{pmp} and~\gls{hjb} branches}\label{sec:connection}
The fourth row of \eref{eq:multishooting_kkt} connects the value function with the Lagrange multipliers associated with the state equations.
By definition, this multiplier corresponds to the next costate value at node $k$, which reveals an interesting connection with the~\gls{pmp} used in indirect multiple shooting methods and the~\gls{kkt} approach, i.e.
\begin{equation*}
    \boldsymbol{\lambda}^+_{k} = \mathcal{V}_{\mathbf{x}_k} + \mathcal{V}_{\mathbf{xx}_k}\delta\mathbf{x}_k.
\end{equation*}
This might be not surprising if we realize that the~\gls{pmp} or~\gls{kkt} approach write the optimal control in term of the costate, while~\gls{hjb} expresses it in terms of the value function.
\eref{eq:multishooting_kkt} is also at the heart of direct single shooting approaches such as~\gls{ddp} if each dynamics gap $\mathbf{\bar{f}}_{k+1}$ vanishes, therefore this connection holds for single shooting settings as well.

Different interpretations can arise from this connection.
For instance, \gls{ddp} or our approach can be interpreted as iterative methods for solving the~\gls{pmp} in discrete-time optimal control problems under single and multiple shooting settings, respectively.
Furthermore, under the context of linear dynamics and quadratic cost, \gls{ddp} or our approach can be classified as global methods as they compute an optimal policy (i.e., a closed-loop solution).

\subsection{Differential dynamic programming with control limits}\label{sec:boxddp}
As proposed by \cite{tassa-icra14}, \textsc{Box-\gls{ddp}} locally approximates the value function at node $k$ as
\begin{equation}
    \label{eq:bounded_bellman}
    \begin{aligned}
        \mathcal{V}_k(\delta\mathbf{x}_k) & = \min\limits_{\delta\mathbf{u}_k} \ell_k(\delta\mathbf{x}_k,\delta\mathbf{u}_k) + \mathcal{V}_{k+1}(\mathbf{f}(\delta\mathbf{x}_k,\delta\mathbf{u}_k)),\\
        \hspace{-2em}\textrm{s.t.} & \hspace{4em} \mathbf{\underline{u}} \leq \mathbf{u}_k + \delta\mathbf{u}_k \leq \mathbf{\bar{u}}~,
    \end{aligned}
\end{equation}
which breaks the optimal control problem into a sequence of simpler sub-problems with control bounds. Then, a local search direction is computed through a \gls{lq} approximation of the value function:
\begin{eqnarray}
    \label{eq:bounded_hamiltonian}
	&&\hspace{-2em}\delta\mathbf{u}^*_k(\delta\mathbf{x}_k) = \nonumber\\\nonumber
	&&\arg\min_{\delta\mathbf{u}_k} \overbrace{\frac{1}{2}
	\begin{bmatrix}
		1 \\ \delta\mathbf{x}_k \\ \delta\mathbf{u}_k
	\end{bmatrix}^\top
	\begin{bmatrix}
		0 & \mathbf{Q}^\top_{\mathbf{x}_k} & \mathbf{Q}^\top_{\mathbf{u}_k} \\
		\mathbf{Q}_{\mathbf{x}_k} & \mathbf{Q}_{\mathbf{xx}_k} & \mathbf{Q}_{\mathbf{xu}_k} \\
		\mathbf{Q}_{\mathbf{u}_k} & \mathbf{Q}^\top_{\mathbf{xu}_k} & \mathbf{Q}_{\mathbf{uu}_k}
	\end{bmatrix}
	\begin{bmatrix}
		1 \\ \delta\mathbf{x}_k \\ \delta\mathbf{u}_k
    \end{bmatrix}}^{Q_k(\delta\mathbf{x}_k,\delta\mathbf{u}_k,\bar{\mathcal{V}}_k)},\\
    &&\hspace{-2em}\textrm{s.t.} \hspace{2em}\mathbf{\underline{u}} \leq \mathbf{u}_k + \delta\mathbf{u}_k \leq \mathbf{\bar{u}},
\end{eqnarray}
where the $\mathbf{Q}_k$ terms describe the \gls{lq} approximation of the action-value function $Q_k(\cdot)$ that can be seen as a function of the derivatives of the value function $\bar{\mathcal{V}}_k=(\mathcal{V}_{\mathbf{x}_k},\mathcal{V}_{\mathbf{xx}_k})$.
Solving~\eref{eq:bounded_hamiltonian} results in a local feedback control law $\delta\mathbf{u}_k=\mathbf{k}_k+\mathbf{K}_k\delta\mathbf{x}_k$ consisting of a feed-forward term $\mathbf{k}_k$ and a state feedback gain $\mathbf{K}_k$ for each discretization point $k$.
Below, we describe how to compute the $\mathbf{Q}_k$ terms in the so-called Riccati sweep step.

\subsubsection{Riccati sweep}\label{sec:riccati_sweep}
The \gls{lq} approximation of the action-value function $Q_k$ is computed recursively, backwards in time, as follows
\begin{eqnarray}
    \label{eq:standard_hamiltonian_computation}\nonumber
	\mathbf{Q}_{\mathbf{x}_k} & = & \boldsymbol{\ell}_{\mathbf{x}_k} + \mathbf{f}^\top_{\mathbf{x}_k}\mathcal{V}_{\mathbf{x}_{k+1}}, \\\nonumber
	\mathbf{Q}_{\mathbf{u}_k} & = & \boldsymbol{\ell}_{\mathbf{u}_k} + \mathbf{f}^\top_{\mathbf{u}_k}\mathcal{V}_{\mathbf{x}_{k+1}}, \\
	\mathbf{Q}_{\mathbf{xx}_k} & = & \boldsymbol{\ell}_{\mathbf{xx}_k} + 
	\mathbf{f}^\top_{\mathbf{x}_k}\mathcal{V}_{\mathbf{xx}_{k+1}} \mathbf{f}_{\mathbf{x}_k},\\\nonumber
	\mathbf{Q}_{\mathbf{xu}_k} & = & \boldsymbol{\ell}_{\mathbf{xu}_k} + 
	\mathbf{f}^\top_{\mathbf{x}_k}\mathcal{V}_{\mathbf{xx}_{k+1}} \mathbf{f}_{\mathbf{u}_k},\\\nonumber
	\mathbf{Q}_{\mathbf{uu}_k} & = & \boldsymbol{\ell}_{\mathbf{uu}_k} + 
	\mathbf{f}^\top_{\mathbf{u}_k}\mathcal{V}_{\mathbf{xx}_{k+1}} \mathbf{f}_{\mathbf{u}_k},
\end{eqnarray}
where $\mathcal{V}_{\mathbf{x}_{k+1}}$, $\mathcal{V}_{\mathbf{xx}_{k+1}}$ are obtained by solving the following algebraic Riccati equations at $k+1$:
\begin{eqnarray}
    \label{eq:riccati_equations}
    \mathcal{V}_{\mathbf{x}_{k+1}} &=& \mathbf{Q}_{\mathbf{x}_{k+1}} - \mathbf{Q}_{\mathbf{xu}_{k+1}} \mathbf{\hat{Q}}_{\mathbf{uu},\text{f}_{k+1}}^{-1} \mathbf{Q}_{\mathbf{u}_{k+1}},\\\nonumber
    \mathcal{V}_{\mathbf{xx}_{k+1}} &=& \mathbf{Q}_{\mathbf{xx}_{k+1}} - \mathbf{Q}_{\mathbf{xu}_{k+1}} \mathbf{\hat{Q}}_{\mathbf{uu},\text{f}_{k+1}}^{-1} \mathbf{Q}_{\mathbf{xu}_{k+1}}^\top,
\end{eqnarray}
with $\mathbf{\hat{Q}}_{\mathbf{uu},\text{f}_{k+1}}$ as the control Hessian in the free space, which we will describe below.
Additionally, we use the gradient and Hessian of the value function to find a local search direction as described below.
In the case of \textsc{Box-\gls{ddp}$^+$}, our adaptation of \textsc{Box-DDP} to allow initialization with state trajectories, the gradient of the value function in the first iteration is relinearized by the \textit{initialization infeasibility} $\mathbf{\bar{f}}^0_{k+1}$ as $\mathcal{V}_{\mathbf{x}_{k+1}}+\mathcal{V}_{\mathbf{xx}_{k+1}}\mathbf{\bar{f}}^0_{k+1}$.

\subsubsection{Control-bounded direction}\label{sec:boxddp_direction}
We compute the control-bounded direction, defined in~\eref{eq:bounded_hamiltonian}, by breaking it down into the so-called \textit{feed-forward} and \textit{feedback} sub-problems~\cite{tassa-icra14}.
We first compute the feed-forward term by solving the following~\gls{qp} program with box constraints:
\begin{equation}
    \label{eq:feedforward_subproblem}
    \begin{aligned}
        \mathbf{k}_k &= \arg\min_{\delta\mathbf{u}_k} \frac{1}{2}\delta\mathbf{u}_k^\top\mathbf{Q}_{\mathbf{uu}_k}\delta\mathbf{u}_k + \mathbf{Q}_{\mathbf{u}_k}^\top\delta\mathbf{u}_k,\\
        \hspace{-2em}\textrm{s.t.} & \hspace{4em} \mathbf{\underline{u}} \leq \mathbf{u}_k + \delta\mathbf{u}_k \leq \mathbf{\bar{u}},
    \end{aligned}
\end{equation}
and then the feedback gain as
\begin{equation}
    \label{eq:feedback_subproblem}
    \mathbf{K}_k=-\mathbf{\hat{Q}}^{-1}_{\mathbf{uu},\text{f}_k}\mathbf{Q}_{\mathbf{ux}_k},
\end{equation}
where $\mathbf{\hat{Q}}_{\mathbf{uu},\text{f}_k}$ is the control Hessian of the free subspace obtained in~\eref{eq:feedforward_subproblem} in which $\text{f}_k$ describes the free subspace at node $k$, i.e., the indexes of the inactive bounds.
With these indexes, we sort and partition the control Hessian as:
\begin{equation}
    \label{eq:control_hessian}
    \mathbf{Q}_{\mathbf{uu}_k}=
    \begin{bmatrix}
    \mathbf{Q}_{\mathbf{uu},\text{f}_k} & \mathbf{Q}_{\mathbf{uu},\text{fc}_k} \\
    \mathbf{Q}_{\mathbf{uu},\text{cf}_k} & \mathbf{Q}_{\mathbf{uu},\text{c}_k}
    \end{bmatrix},
\end{equation}
and compute $\mathbf{\hat{Q}}^{-1}_{\mathbf{uu},\text{f}_k}$ internally based on the factorization of $\mathbf{Q}^{-1}_{\mathbf{uu},\text{f}_k}$.
This is what our~~\textsc{Box-\gls{qp}} program does to solve the feed-forward sub-problem efficiently via the Projected-Newton~\gls{qp} algorithm~\cite{bertsekas-siam82}.
This algorithm quickly identifies the active set and moves along the free subspace of the Newton step.
It also has a similar computational cost to the unconstrained \gls{qp} if the active set remains unchanged.
Thus, the runtime performance is similar to the~\gls{ddp} algorithm.
However, it requires a feasible initialization $\delta\mathbf{u}^0_k$.

Again, by using a Projected-Newton~\gls{qp} algorithm, we further efficiently obtain the control Hessian of the free subspace $\mathbf{Q}^{-1}_{\mathbf{uu},\text{f}_k}$ as the algorithm computes it internally when it moves along the free subspace of the Newton step.
With it, we compute a state feedback gain that generates corrections within the control limits.
This is an important feature for controlling the robot as well as for rolling-out the nonlinear dynamics in the forward pass.
For more details about the Projected-Newton QP algorithm see~\cite{bertsekas-siam82}.

\subsubsection{State integration}
In any~\gls{ddp} algorithm such as~\textsc{Box-\gls{ddp}}, we perform a state integration using the locally-linear policy as
\begin{eqnarray}
	\mathbf{\hat{x}}_0 &=& \mathbf{x}_0,\\\nonumber
	\mathbf{\hat{u}}_k &=& \mathbf{u}_k + \alpha \mathbf{k}_k + \mathbf{K}_k (\mathbf{\hat{x}}_k \ominus \mathbf{x}_k),\hspace*{1em}\scriptstyle{\forall k=\{0,1,\cdots,N-1\}}\\\nonumber
	\mathbf{\hat{x}}_{k+1} &=& \mathbf{f}(\mathbf{\hat{x}}_k, \mathbf{\hat{u}}_k),\hspace*{8em}\scriptstyle{\forall k=\{0,1,\cdots,N-1\}}
\end{eqnarray}
where $\mathbf{\hat{x}}_k$, $\mathbf{\hat{u}}_k$ are the new state and control at node $k$ generated using a step length $\alpha$.
The feedback gain helps to distribute the nonlinearities; however, as seen in the previous section, it does not resemble the numerical behavior described by the~\gls{fonc} of optimality in direct multiple shooting.
This different numerical behavior stems from the state integration procedure closing the gaps, i.e., $\mathbf{\bar{f}}_k = \mathbf{0}$, $\forall k=\{0,1,\cdots,N\}$.

\subsubsection{Expected improvement}
When solving the algebraic Riccati equations, we obtain the expected improvement as
\begin{equation}\label{eq:costate_rate}
	\Delta \mathcal{V}_k = -\frac{1}{2}\mathbf{Q}_{\mathbf{u}_k}^T \mathbf{\hat{Q}}_{\mathbf{uu},\text{f}_k}^{-1} \mathbf{Q}_{\mathbf{u}_k}.
\end{equation}
Below, we elaborate the proposed algorithm based on the aforementioned description.

\section{Box-FDDP: a feasibility-driven approach for multiple shooting}\label{sec:boxfddp}
We now introduce a novel algorithm that combines a feasibility-driven search (\sref{sec:direct_transcription}) with an active set treatment of the control limits (\sref{sec:boxddp}) named \textsc{Box-\acrshort{fddp}}.
The \textsc{Box-\acrshort{fddp}} algorithm comprises two modes: \textit{feasibility-driven} and \textit{control-bounded} modes, one of which is chosen for a given iteration~(\aref{alg:boxfddp}).
The feasibility-driven\footnote{Here, \textit{feasibility} concerns the dynamics of the system, not the feasibility of other problem constraints.} mode mimics the numerical resolution of a direct multiple
shooting problem with only dynamics constraints when computing the search direction and step length (lines \ref{alg:search_direction_start}-\ref{alg:search_direction_end} and \ref{alg:line_search}-\ref{alg:step_length_end}, respectively).
This mode neglects the control limits of the system as its focuses on dynamic feasibility only.
In contrast, the control-bounded mode projects the search direction onto the feasible control region whenever the dynamics constraint is feasible (line \ref{alg:control_bounded_direction}).
In both modes, the applied controls in the forward pass are projected onto their feasible box (line \ref{alg:project_control}), causing dynamically-infeasible iterations to reach the control box.
With this strategy, our solver focuses on feasibility early on, which increases its basins of attraction, and later on optimality.
Technical descriptions of both modes are elaborated in~Sections~\ref{sec:search_direction}~and~\ref{sec:step_length}.
Note that the existence of feasible descent directions are introduced later in~\sref{sec:formal_guarantees}.

\begin{algorithm}%
    \BlankLine
    compute \acrshort{lq} approximation of the cost and dynamics\\ \label{alg:lq_appr_problem}
    \If{infeasible iterate}{
        compute the gaps,~\eref{eq:gap}\\ \label{alg:compute_gaps}
    }
    \For{$k\leftarrow N-1$ \KwTo $0$}{ \label{alg:search_direction_start}
        update the feasibility-driven $Q-$function,~\eref{eq:hamiltonian_computation}\\ \label{alg:update_hamiltonian}
        \eIf{infeasible iterate}{
            compute feasibility-driven direction,~\eref{eq:feedforward_feedback_feasibility}\\ \label{alg:control_free_direction}
        }{
            project \textsc{Box-\gls{qp}} warm start,~\eref{eq:clamp_warmstart}\\ \label{alg:clamp_warmstart}
            compute control-bounded direction,~\eref{eq:feedforward_subproblem}-(\ref{eq:feedback_subproblem})\\ \label{alg:control_bounded_direction}
        }
    }\label{alg:search_direction_end}
    \For{$\alpha \in \left\{1, \frac{1}{2}, \cdots, \frac{1}{2^n}\right\}$}{\label{alg:line_search}
        \For{$k\leftarrow 0$ \KwTo $N$}{
            project control onto the feasible box,~\eref{eq:clamp_rollout}\\ \label{alg:project_control}
            \eIf{infeasible iterate or $\alpha \neq 1$}{
                update the gaps,~\eref{eq:gaps_dynamics}\\ \label{alg:update_gaps}
            }{
                close the gaps, $\mathbf{f}_k=\mathbf{0}\hspace{0.5em}\forall k\in\{0, \cdots, N-1\}$\\ \label{alg:close_gaps}
            }
            perform step,~\eref{eq:nonlinear_step}\\ \label{alg:perform_step}
        }
        compute the expected improvement,~\eref{eq:expected_model}\\ \label{alg:compute_expected_model}
        \If{success step}{break}\label{alg:success_step}
    }\label{alg:step_length_end}
    \caption{Control-limited \acrshort{fddp} (\textsc{Box-\acrshort{fddp}})}
    \label{alg:boxfddp}
\end{algorithm}

\subsection{Search direction}\label{sec:search_direction}
In the standard \textsc{Box-\gls{ddp}} algorithm, an initial forward pass is performed to obtain the initial state trajectory $\mathbf{x}_s$.
This trajectory enforces the dynamics explicitly; thus, the gaps are zero, i.e., $\mathbf{\bar{f}}_{k}=\mathbf{0}$ for all $k=\{0,1,\cdots,N-1\}$.
Instead, our multiple shooting variant, \textsc{Box-\gls{fddp}}, computes the gaps once at each iteration (line~\ref{alg:compute_gaps}), which are used to find the search direction and to compute the expected improvement.
However, if the iteration is dynamically feasible, then the search direction procedure is the same as in the standard \textsc{Box-\gls{ddp}}~\cite{tassa-icra14}.
Below, we describe the steps performed to compute the search direction.

\subsubsection{Computing the gaps}\label{sec:gaps}
Given a current iterate $(\mathbf{x}_s,\mathbf{u}_s)$, we perform a nonlinear roll-out to compute the gaps as
\begin{equation}\label{eq:gap}
	\mathbf{\bar{f}}_{k+1} := \mathbf{f}(\mathbf{x}_k,\mathbf{u}_k) \ominus \mathbf{x}_{k+1}, \hspace*{1em} \scriptstyle{\forall k=\{0,1,\cdots,N-1\}}
\end{equation}
where $\mathbf{f}(\mathbf{x}_k,\mathbf{u}_k)$ is the roll-out state at interval $k+1$, $\mathbf{x}_{k+1}$ is the next shooting state, and $\ominus$ is the difference operator.

\subsubsection{Action-value function of direct multiple shooting formulation}
Without loss of generality, we use the \acrfull{gn} approximation~\cite{li-icinco04} to write the action-value function of our algorithm as
\begin{equation}
    \label{eq:hamiltonian_standard_form}
    \begin{aligned}
    Q_k(\cdot)&=
    \frac{1}{2}
	\begin{bmatrix}
		1 \\ \delta\mathbf{x}_{k+1}
	\end{bmatrix}^\top
	\begin{bmatrix}
		0 & \mathcal{V}^\top_{\mathbf{x}_{k+1}} \\
		\mathcal{V}_{\mathbf{x}_{k+1}} & \mathcal{V}_{\mathbf{xx}_{k+1}}
	\end{bmatrix}
	\begin{bmatrix}
		1 \\ \delta\mathbf{x}_{k+1}
    \end{bmatrix}&\\
    &+
    \frac{1}{2}
	\begin{bmatrix}
		1 \\ \delta\mathbf{x}_k \\ \delta\mathbf{u}_k
	\end{bmatrix}^\top
	\begin{bmatrix}
		0 & \boldsymbol{\ell}^\top_{\mathbf{x}_k} & \boldsymbol{\ell}^\top_{\mathbf{u}_k} \\
		\boldsymbol{\ell}_{\mathbf{x}_k} & \boldsymbol{\ell}_{\mathbf{xx}_k} & \boldsymbol{\ell}_{\mathbf{xu}_k} \\
		\boldsymbol{\ell}_{\mathbf{u}_k} & \boldsymbol{\ell}^\top_{\mathbf{xu}_k} & \boldsymbol{\ell}_{\mathbf{uu}_k}
	\end{bmatrix}
	\begin{bmatrix}
		1 \\ \delta\mathbf{x}_k \\ \delta\mathbf{u}_k
    \end{bmatrix}&
    \end{aligned},
\end{equation}
where again $\boldsymbol{\ell}_{\mathbf{x}_k},\boldsymbol{\ell}_{\mathbf{u}_k}$ and $\boldsymbol{\ell}_{\mathbf{xx}_k},\boldsymbol{\ell}_{\mathbf{xu}_k},\boldsymbol{\ell}_{\mathbf{uu}_k}$ describe the gradient and Hessian of the cost function, respectively; $\delta\mathbf{x}_{k+1}=\mathbf{f}_{\mathbf{x}_k}\delta\mathbf{x}_k+\mathbf{f}_{\mathbf{u}_k}\delta\mathbf{u}_k$ is the linearized dynamics; and $\mathbf{f}_{\mathbf{x}_k},\mathbf{f}_{\mathbf{u}_k}$ are its Jacobians.

In direct multiple shooting settings, linearization of the system dynamics includes a drift term
\begin{equation}\label{eq:gaps}
    \delta\mathbf{x}_{k+1} = \mathbf{f_x}_k\delta\mathbf{x}_k + \mathbf{f_u}_k\delta\mathbf{u}_k + \mathbf{\bar{f}}_{k+1},
\end{equation}
as there are gaps in the dynamics term $\mathbf{\bar{f}}_{k+1}$ produced between subsequent shooting segments (q.v.~\fref{fig:multiple_shooting}).
Then, the Riccati sweep needs to be adapted as follows:
\begin{eqnarray}
    \label{eq:hamiltonian_computation}\nonumber
	\mathbf{Q}_{\mathbf{x}_k} & = & \boldsymbol{\ell}_{\mathbf{x}_k} + \mathbf{f}^\top_{\mathbf{x}_k}\mathcal{V}^+_{\mathbf{x}_{k+1}}, \\\nonumber
	\mathbf{Q}_{\mathbf{u}_k} & = & \boldsymbol{\ell}_{\mathbf{u}_k} + \mathbf{f}^\top_{\mathbf{u}_k}\mathcal{V}^+_{\mathbf{x}_{k+1}}, \\
	\mathbf{Q}_{\mathbf{xx}_k} & = & \boldsymbol{\ell}_{\mathbf{xx}_k} + 
	\mathbf{f}^\top_{\mathbf{x}_k}\mathcal{V}_{\mathbf{xx}_{k+1}} \mathbf{f}_{\mathbf{x}_k},\\\nonumber
	\mathbf{Q}_{\mathbf{xu}_k} & = & \boldsymbol{\ell}_{\mathbf{xu}_k} + 
	\mathbf{f}^\top_{\mathbf{x}_k}\mathcal{V}_{\mathbf{xx}_{k+1}} \mathbf{f}_{\mathbf{u}_k},\\\nonumber
	\mathbf{Q}_{\mathbf{uu}_k} & = & \boldsymbol{\ell}_{\mathbf{uu}_k} + 
	\mathbf{f}^\top_{\mathbf{u}_k}\mathcal{V}_{\mathbf{xx}_{k+1}} \mathbf{f}_{\mathbf{u}_k}
\end{eqnarray}
in which
\begin{equation}
    \label{eq:value_computation}
    \mathcal{V}^+_{\mathbf{x}_{k+1}}=\mathcal{V}_{\mathbf{x}_{k+1}}+\mathcal{V}_{\mathbf{xx}_{k+1}}\mathbf{\bar{f}}_{k+1}
\end{equation}
is the gradient of the value function after the deflection produced by $\mathbf{\bar{f}}_{k+1}$ (also described above as relinearization).
Note that the Hessian of the value function remains unchanged as \gls{ddp} approximates the value function through a \gls{lq} model.
Additionally, this modification affects the values of the Riccati equations,~\eref{eq:riccati_equations}, and expected improvement,~\eref{eq:costate_rate}, as they depend on the gradient of the value function.

\subsubsection{Feasibility-driven direction}\label{sec:free_direction}
For \textit{dynamically-infeasible} iterates (line~\ref{alg:control_free_direction}), we ignore the control constraints and compute a \textit{control-unbounded} direction:
\begin{equation}
    \label{eq:feedforward_feedback_feasibility}
    \begin{aligned}
    \mathbf{k}_k &=-\mathbf{Q}_{\mathbf{uu}_k}^{-1}\mathbf{Q}_{\mathbf{u}_k},&\\
    \mathbf{K}_k &=-\mathbf{Q}_{\mathbf{uu}_k}^{-1}\mathbf{Q}_{\mathbf{ux}_k}.&
    \end{aligned}
\end{equation}
We do this because we cannot quantify the effect of the gaps on the control bounds, which are needed to solve the feed-forward sub-problem~\eref{eq:feedforward_subproblem}.
Our approach is equivalent to opening the control bounds during dynamically-infeasible iterates.

\subsubsection{Control-bounded direction}\label{sec:bounded_direction}
We warm-start the \textsc{Box-\gls{qp}} using the feed-forward term $\mathbf{k}_k$ computed in the previous iteration.
However, if the algorithm is switching from feasibility to control-bounded mode (i.e., the previous iteration is infeasible), then $\mathbf{k}_k$ might fall outside the feasible box and  $\mathbf{\underline{u}}-\mathbf{u}_k\leq\mathbf{k}_k\leq\mathbf{\bar{u}}-\mathbf{u}_k$ do not hold.
This violates the assumption of the previously-described \textsc{Box-\gls{qp}}, for which a feasible initial point needs to be provided.

To handle infeasible iterates, we propose to project the warm-start of the \textsc{Box-\gls{qp}} (line~\ref{alg:clamp_warmstart}) as
\begin{equation}
    \label{eq:clamp_warmstart}
    \llbracket\mathbf{k}_k\rrbracket_{\mathbf{\underline{u},\bar{u}}} = \min{(\max{(\mathbf{k}_k,\mathbf{\underline{u}}-\mathbf{u}_k)},\mathbf{\bar{u}}-\mathbf{u}_k)},
\end{equation}
where $\mathbf{\underline{u}}$, $\mathbf{\bar{u}}$ are the lower and upper bounds of the feed-forward sub-problem,~\eref{eq:feedforward_subproblem}, respectively.

Once we project the warm-start $\mathbf{k}_k$, we solve the feed-forward and feedback sub-problems as explained in~\sref{sec:boxddp_direction}.
Furthermore, we solve the \textsc{Box-\gls{qp}} using a Projected-Newton method~\cite{bertsekas-siam82}, which handles box constraints efficiently as described above.

\subsection{Step length}\label{sec:step_length}
As far as we know, the standard \textsc{Box-\gls{ddp}}~\cite{tassa-icra14} modifies only the search direction (i.e., backward pass) to handle the control limits.
However, it is also important to project the controls onto the feasible box during the forward pass.
We do this by finding a step length that minimizes the cost~\cite{nocedal-optbook}.

\subsubsection{Projecting the roll-out towards the feasible box}
We propose to project the controls onto the feasible box in the nonlinear roll-out (line~\ref{alg:project_control}), i.e.,
\begin{equation}
    \label{eq:clamp_rollout}
    \mathbf{\hat{u}}_k \leftarrow \min{(\max{(\mathbf{\hat{u}}_k,\mathbf{\underline{u}})},\mathbf{\bar{u}})},
\end{equation}
where $\mathbf{\hat{u}}_k$ is the updated control from the control policy.
Our method does not require to solve another \gls{qp} problem~\cite{xie-icra17} or to project the linear search direction given the gaps on the dynamics~\cite{howell-19}.
Furthermore, the control policy considers a gap prediction that guarantees a feasible descent direction.
We formally describe the technical details of this procedure in~\sref{sec:nonlinear_step}.

\subsubsection{Updating the gaps}
As analyzed earlier, the evolution of the gaps in direct multiple shooting is affected by the selected step length.
For an optimal control problem without control limits, this evolution is defined as
\begin{equation}
    \label{eq:gaps_dynamics}
    \mathbf{\bar{f}}_k \leftarrow (1-\alpha)\mathbf{\bar{f}}_k,
\end{equation}
where $\alpha$ is the step-length found by the line-search procedure (line~\ref{alg:line_search}-\ref{alg:success_step}).
Note that a full step $(\alpha=1)$ closes the gaps completely.
We described this gap contraction rate in~\sref{sec:fddp_kkt}.

\subsubsection{Nonlinear step}\label{sec:nonlinear_step}
With a nonlinear roll-out\footnote{In this work, a nonlinear \textit{roll-out} is also referred to as a nonlinear step.} (line~\ref{alg:perform_step}), we avoid the linear prediction error of the dynamics that is typically handled by a \textit{merit} function in general-purpose \gls{nlp} algorithms, as explained in~\cite{mastalli-icra20}.
If we keep the gap-contraction rate of~\eref{eq:gaps_dynamics}, then we obtain
\begin{eqnarray}
    \label{eq:nonlinear_step}
	\mathbf{\hat{x}}_{k} &=& \mathbf{f}(\mathbf{\hat{x}}_{k-1},\mathbf{\hat{u}}_{k-1}) \oplus (\alpha - 1)\mathbf{\bar{f}}_{k-1}, \nonumber\\
	\mathbf{\hat{u}}_k &=& \mathbf{u}_k + \alpha\mathbf{k}_k + \mathbf{K}_k(\mathbf{\hat{x}}_k\ominus\mathbf{x}_k),
\end{eqnarray}
where $\mathbf{\hat{x}}_{k}$, $\mathbf{\hat{u}}_{k}$ are the next state and control along an $\alpha$-step; $\mathbf{k}_k$ and $\mathbf{K}_k$ are the feed-forward term and feedback gains computed by~\eref{eq:feedforward_feedback_feasibility} or~\eref{eq:feedforward_subproblem}-(\ref{eq:feedback_subproblem}).
Furthermore, the initial condition of the roll-out is defined as $\mathbf{\hat{x}}_0=\mathbf{\tilde{x}}_0\oplus(\alpha-1)\mathbf{\bar{f}}_0$.
Note that this is in contrast to the standard \textsc{Box-\gls{ddp}}, in which the gaps are always closed, even for $\alpha < 1$.

Avoiding the use of a merit function helps the algorithm to check the search direction more accurately.
Indeed, it has been shown that the nonlinear roll-out is more effective than a standard line search procedure as it reduces the number of iterations~\cite{liao-92}.

\subsubsection{Expected improvement}\label{sec:expected_model}
It is critical to properly evaluate the success of a trial step.
Given the current dynamics gaps $\bar{\mathbf{f}}_k$, \textsc{Box-\gls{fddp}} computes the expected improvement of a computed search direction as
\begin{equation}\label{eq:expected_model}
	\Delta J(\alpha) = \Delta_1\alpha + \frac{1}{2}\Delta_2\alpha^2,
\end{equation}
with
\begin{eqnarray}
	\Delta_1 = \sum_{k=0}^N \mathbf{k}_k^\top\mathbf{Q}_{\mathbf{u}_k} +\mathbf{\bar{f}}_k^\top(\mathcal{V}_{\mathbf{x}_k} - \mathcal{V}_{\mathbf{xx}_k}\delta\mathbf{\hat{x}}_k),\nonumber\\
	\Delta_2 = \sum_{k=0}^N \mathbf{k}_k^\top\mathbf{\hat{Q}}_{\mathbf{uu},\text{f}_k}\mathbf{k}_k + \mathbf{\bar{f}}_k^\top(2\mathcal{V}_{\mathbf{xx}_k}\delta\mathbf{\hat{x}}_k - \mathcal{V}_{\mathbf{xx}_k}\mathbf{\bar{f}}_k),
\end{eqnarray}
where $\mathbf{\hat{Q}}_{\mathbf{uu},\text{f}_k}$ is the control Hessian of the free space, $\delta\mathbf{\hat{x}}_k = \mathbf{\hat{x}}_k\ominus\mathbf{x}_k$, and $J$ is the total cost of a given state-control trajectory ($\mathbf{x}_s$, $\mathbf{u}_s$).
We use this expected improvement model for both modes.
Note that, in the feasibility-driven mode, the free space spans the entire control space; instead, in the control-bounded mode, the gaps are zero.

We obtain this expression by computing the cost from a linear roll-out of the current control policy as described in~\eref{eq:nonlinear_step}.
We also accept ascent directions when evaluating the trial step, our approach is inspired by the Goldstein condition~\cite[Chapter~3]{nocedal-optbook}:
\begin{equation}
	 \ell'-\ell \le
	\begin{cases}
  b_1 \Delta J(\alpha) & \textrm{if }\Delta J(\alpha)\le 0 \\
  b_2 \Delta J(\alpha) & \textrm{otherwise}
	\end{cases},
\end{equation}
where $b_1$, $b_2$ are adjustable parameters, we used in this paper $b_1 = 0.1$ and $b_2 = 2$.
Ascent directions improve the algorithm convergence as it helps to reduce the feasibility error through a moderate increment in the cost.

\subsubsection{Regularization}
We regularize the $\mathbf{Q_{uu}}$ and $\mathcal{V}_\mathbf{{xx}}$ terms through a Levenberg-Marquardt scheme~\cite{fletcher-71}.
Concretely, we increase the damping value $\mu$ when the computation of the feed-forward sub-problem---formulated in \eref{eq:feedforward_subproblem}---fails, or when the forward pass accepts a step length smaller than $\alpha_0=0.01$.
Moreover, we decrease the damping value if the iteration accepts a step larger than $\alpha_1=0.5$.
Both regularization procedures modify the values of the $\mathbf{Q_{uu}}$ and $\mathcal{V}_\mathbf{{xx}}$ terms during the Riccati sweep computation as
\begin{equation}\nonumber
\begin{aligned}
    \mu' &\leftarrow \beta_{i,d}\mu,\\\nonumber
    \mathbf{Q_{uu}} &\leftarrow \mathbf{Q_{uu}} + \mu'\mathbf{I},\\\nonumber
    \mathcal{V}_\mathbf{{xx}} &\leftarrow \mathcal{V}_\mathbf{{xx}} + \mu'\mathbf{I},
\end{aligned}
\end{equation}
where $\beta_i$ and $\beta_d$ are the factors\footnote{
$\beta_{i,d}$ commonly range between $2$--$10$.
We set $\beta_{i,d} = 10$ in this work.} used to increase or decrease the current damping value $\mu$, respectively;  $\mu'$ is the newly-computed damping value; and $\mathbf{I}$ is the identity matrix.
Additionally, we start the regularization procedure with an initial, and user-defined, damping value.\footnote{We use $10^{-9}$ as the initial regularization value.}
We also define minimum and maximum damping values to avoid increasing or decreasing the damping value unnecessarily.\footnote{We use $10^{-16}$ and $10^{12}$ as the minimum and maximum damping values, respectively. Note that $10^{-16}$ is approximately the resolution of a \texttt{double} number.}

Both regularizations significantly increase the robustness of the algorithm and ensures convergence, as it moves from Newton direction to steepest-descent, and vice versa.
The Newton direction, which occurs with $\mu=0$, provides fast convergence and is robust against scaling because it exploits the Hessian of the problem.
However, it does not always produce valid descent directions as $\mathbf{Q_{uu}}$ might be indefine and the problem nonconvex.
In such cases, increasing the damping value guarantees that $\mathbf{Q_{uu}}$ is positive-define which, in turn, computes a search direction closer to the steepest-descent one.
Instead, $\mathcal{V}_\mathbf{{xx}}$ enforces the state trajectory to be closer to the one previously computed~\cite{tassa-thesis}.
It will also not result in vanishing feedback gains even for large damping values.

\subsection{Existence of feasible descent directions}\label{sec:formal_guarantees}
As described above, our approach has two main modes: feasibility-driven and control-bounded.
During the feasibility-driven phase, we compute a search direction to drive the next guess towards dynamic feasibility and try a step while keeping the control within the box constraints.
This projection procedure can be seen as a nonlinear term in our dynamics, but we assume its effect is negligible for finding a feasible direction.
On the other hand, our algorithm computes a search direction that considers the box constraints after the dynamic feasibility has been achieved.
This is needed to improve the next current guess by taking control constraints into account when computing the feedback gains along the free subspace.

As analyzed in~\sref{sec:fddp_kkt}, the feasibility-driven direction is computed by mimicking the numerical behavior of a nonlinear program during the resolution of a direct multiple shooting problem with only dynamics constraints.
It implies that the feasibility-driven search produces a descent direction, and eventually the algorithm converges, if the cost Hessian is a positive definite matrix.
Indeed, the positiveness is always guaranteed by our regularization procedure as described before.
Furthermore, the feasibility-driven step aims at reducing the nonlinearities produced by a single shooting formulation (e.g.,~\gls{ddp} algorithm).
When the dynamics are feasible, we apply a control-bounded search which also produces a descent direction as it boils down to a~\gls{qp} program.

In the next section, we present a series of results that demonstrate the benefits of our feasibility-driven approach.
\section{Results}\label{sec:results}
We analyze \textsc{Box-\gls{fddp}} across a wide range of optimal control problems (briefly introduced in~\sref{sec:problems}) as follows.
First, in~\sref{sec:feasibility_advantages}, we show the benefits of the feasibility mode by analyzing the gap contraction and its connection with the nonlinearities in the dynamics.
Second, we compare our algorithm against a direct transcription formulation in~\sref{sec:knitro_comparison}.
Concretely, we compare the dynamic feasibility and optimality evolutions, runtime performance, and robustness to different initial guesses against the interior point and active set algorithms available in \textsc{Knitro}.
Finally, in~\sref{sec:squashing_analisis}, we report the results of a squashing approach for solving the control bounds as it demonstrates the numerical performance of having two modes.

\begin{figure*}%
    \centering\begin{tabular}{cc}
    \rowname{a} & \href{https://youtu.be/bOGBPTh_lsU?t=18}{\raisebox{-.5\height}{\includegraphics[width=0.93\textwidth]{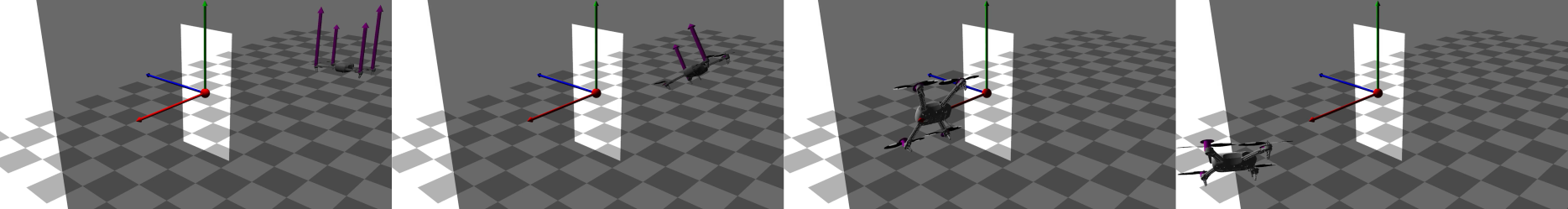}}}\\
    & \rule[0.5em]{16.9cm}{0.1pt}\\
    \rowname{b} & \href{https://youtu.be/bOGBPTh_lsU?t=8}{\raisebox{-.5\height}{\includegraphics[width=0.93\textwidth]{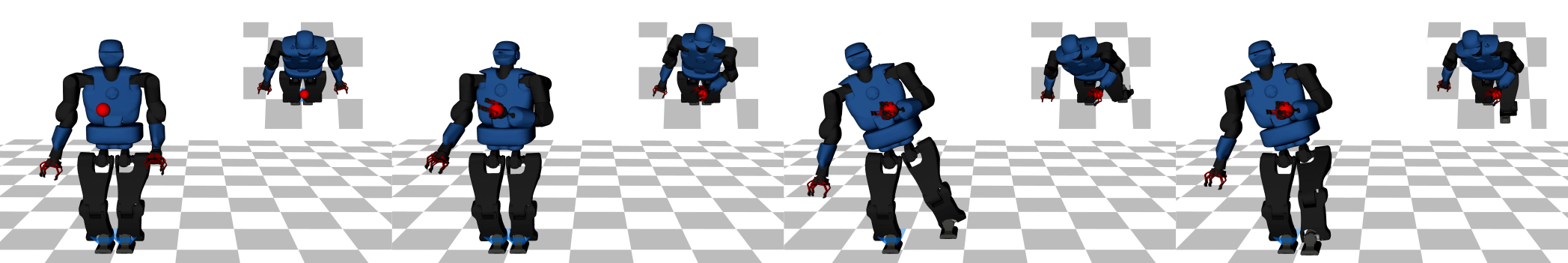}}}\\
    & \rule[0.5em]{16.9cm}{0.1pt}\\
    \rowname{c} & \href{https://youtu.be/bOGBPTh_lsU?t=60}{\raisebox{-.5\height}{\includegraphics[width=0.93\textwidth]{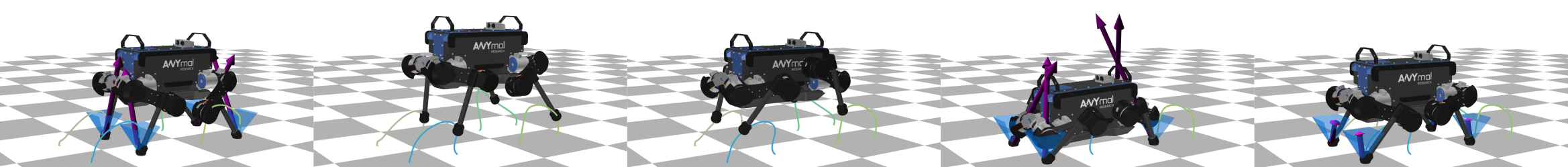}}}\\
    & \rule[0.5em]{16.5cm}{0.1pt}\\
    \rowname{d} & \href{https://youtu.be/bOGBPTh_lsU?t=42}{\raisebox{-.5\height}{\includegraphics[width=0.93\textwidth]{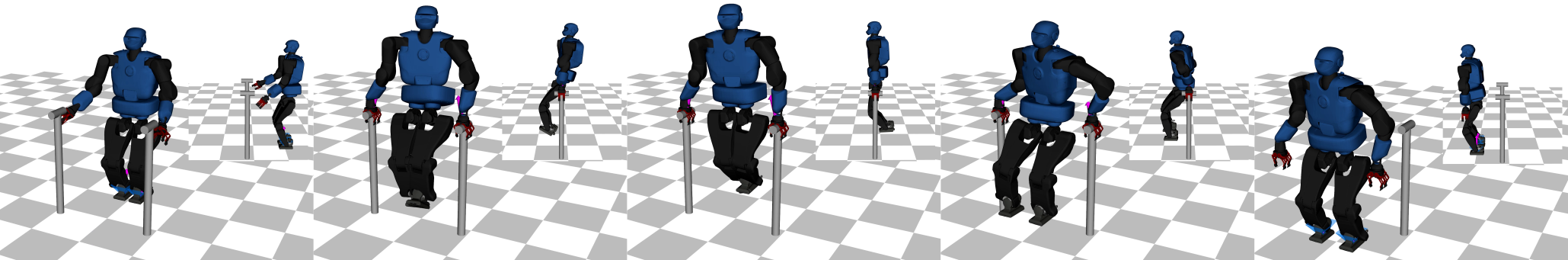}}}\\
    & \rule[0.5em]{16.5cm}{0.1pt}\\
    \rowname{e} & \href{https://youtu.be/bOGBPTh_lsU?t=85}{\raisebox{-.5\height}{\includegraphics[width=0.93\textwidth]{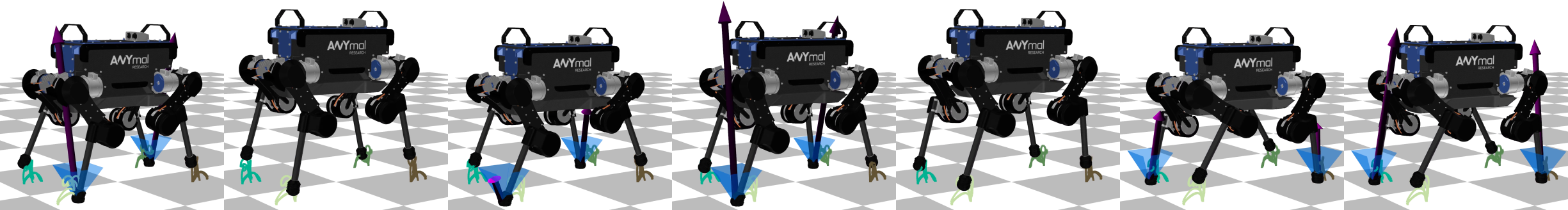}}}\\
    & \rule[0.5em]{16.5cm}{0.1pt}\\
    \rowname{f} & \href{https://youtu.be/bOGBPTh_lsU?t=106}{\raisebox{-.5\height}{\includegraphics[width=0.93\textwidth]{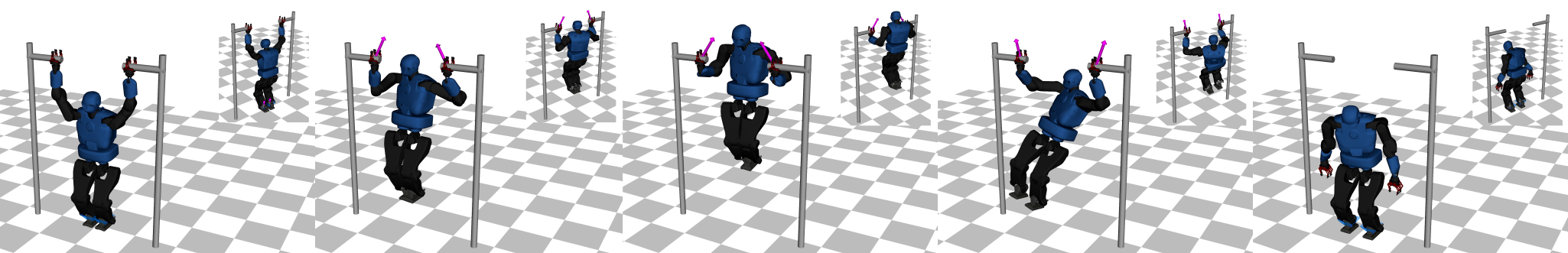}}}
    \end{tabular}
    \caption{Snapshots of different robot maneuvers computed by \textsc{Box-\gls{fddp}}.
    (a) traversing a narrow passage with a quadcopter (\textsc{goal}); (b) Talos balancing on a single leg (\textsc{taichi}); (c) aggressive jumping of 30~\si{\centi\metre} that reaches ANYmal limits (\textsc{jump}); (d) Talos dipping on a parallel bars (\textsc{dip}); (e) ANYmal hopping with two legs (\textsc{hop}); (f) Talos performing a pull-up bar task (\textsc{pullup}).
    To watch the video, click its respective figure or see \texttt{\url{https://youtu.be/bOGBPTh_lsU}}.}
    \label{fig:highly_dyn_maneuvers}
\end{figure*}

We benchmark the algorithms using an Intel Core i9-9900KF CPU with eight cores @ 3.60GHz and 16 MB cache.
Our implementation of \textsc{Box-\gls{fddp}} supports code generation, but all runtime performances reported henceforth do not employ it for fair comparison with other approaches.
We use 8 threads to compute the cost and dynamics derivatives for the experiments with the \textsc{Box-\gls{fddp}} algorithm and the legged robots only.
We use the same initial regularization and stopping criteria values for each problem, and the values are $10^{-9}$ and $5\times 10^{-5}$, respectively.

\subsection{Optimal control problems}\label{sec:problems}
To provide empirical evidence on the benefits of the feasibility-driven approach, we developed a range of different optimal control problems: 1) an under-actuated double pendulum (\textsc{pend}); 2) a quadcopter navigating towards a goal (\textsc{goal}) or through a narrow passage (\textsc{narrow}) and looping (\textsc{loop}); 3) various gaits, aggressive jumps (\textsc{jump}) and unstable hopping (\textsc{hop}) in a quadruped robot; 4) whole-body manipulation (\textsc{man}), hand control while balancing in single leg (\textsc{taichi}), dip on parallel bars (\textsc{dip}) and a pull-up bar task (\textsc{pullup}) in a humanoid robot.
\fref{fig:highly_dyn_maneuvers} shows snapshots of motions computed by \textsc{Box-\gls{fddp}} for some of these problems, and the accompanying video shows the entire motion sequences.\footnote{Supplementary video:~\texttt{\url{https://youtu.be/bOGBPTh_lsU}}.}

We describe the cost functions, dynamics, control limits, penalization terms, and initialization of each optimal control problem in Appendix~\ref{sec:oc_problems}.
Finally, some of these problems, as well as our implementation of the \textsc{Box-\gls{fddp}} algorithm, are publicly available in the~\textsc{Crocoddyl} repository~\cite{crocoddylweb}.

\subsection{Advantages of the feasibility-driven mode}\label{sec:feasibility_advantages}
To understand the benefits of the feasibility-driven mode, we analyze the resulting total cost, number of iterations and total computation time obtained in both algorithms: \textsc{Box-\gls{fddp}} and \textsc{Box-\gls{ddp}}$^+$ using the same initial guess.
\textsc{Box-\gls{ddp}}$^+$ is an improved version of the standard \textsc{Box-\gls{ddp}} proposed in~\cite{tassa-icra14}, which it accepts initialization for both: state and control trajectories as described above.
This version accepts infeasible warm-starts as in~\eref{eq:hamiltonian_computation}, and it is available in the \textsc{Crocoddyl} repository.
Without this modification, the standard \textsc{Box-\gls{ddp}} could easily diverge (and not converge at all) when we initialize it using quasi-static torques in problems with medium to longer horizons.

\subsubsection{Larger basin of attraction and convergence}
In our experiments, \textsc{Box-\gls{ddp}}$^+$ was unable to generate jumping and hopping motions for quadrupeds, as well as pull-ups for humanoids, i.e., it failed to solve our \textsc{jump}, \textsc{hop}, and \textsc{pullup} task specifications (marked by the \xmark~in \tref{tab:results}).
\textsc{Box-\gls{ddp}}$^+$ failed to generate such aggressive motions because trajectories satisfying all the specifications of those tasks are significantly distant from the initial guess provided to the solver.
\textsc{Box-\gls{ddp}}$^+$ behaves poorly as it computes solution with higher cost value and computation time, which is critical drawback in model predictive control applications.

\begin{figure}%
  \centering
  \includegraphics[width=\linewidth]{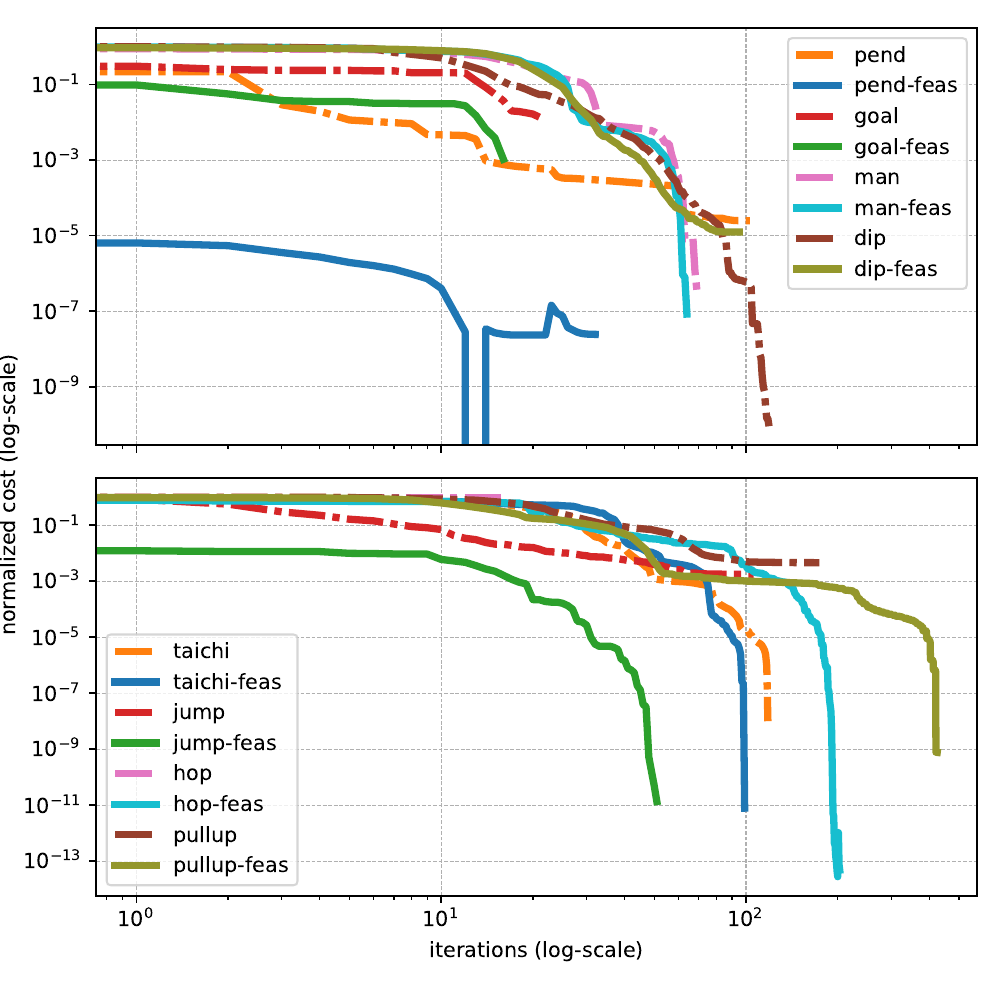}\vspace{-0.8em}
  \caption{Cost and convergence comparison for different optimal control problems.
  \textsc{Box-\gls{fddp}} outperformed \textsc{Box-\gls{ddp}}$^+$ in all the cases: (top) double pendulum (\textsc{pend}), quadcopter navigation (\textsc{quad}), and whole-body manipulation (\textsc{man}), Talos dipping on parallel bars (\textsc{dip}); and (bottom) whole-body balance (\textsc{taichi}), quadrupedal jumping (\textsc{jump}), quadrupedal hopping (\textsc{hop}), Talos' pull-up workout (\textsc{pullup}).
  \textsc{Box-\gls{fddp}} (*-feas) solved the problem with fewer iterations and often with lower cost than \textsc{Box-\gls{ddp}}$^+$.
  Furthermore, \textsc{Box-\gls{ddp}}$^+$ failed to solve some of the hardest problems: i.e., quadrupedal jumping and hopping, Talos' pull-up task.
  Our algorithm showed a large basin of attraction for local optimum as it is less sensitive to poor initialization compared with \textsc{Box-\gls{ddp}}$^+$.
  We use the same initial guess for both cases: \textsc{Box-\gls{fddp}} (*-feas) and \textsc{Box-\gls{ddp}}$^+$.}
  \label{fig:cost_iterations}
\end{figure}

In contrast, our approach (\textsc{Box-\gls{fddp}}) was able to solve all of the tasks, and it did so with fewer iterations and lower total cost (see \tref{tab:results} and \fref{fig:cost_iterations}).
Furthermore, \textsc{Box-\gls{fddp}} and \textsc{Box-\gls{ddp}}$^+$ have the same algorithmic complexity, but since our approach requires fewer iterations than \textsc{Box-\gls{ddp}}$^+$, the total computation time of our approach is lower.
These results are a direct consequence of the feasibility-driven mode in our approach, which is able to find control sequences even when faced with poor initial guesses.
The infeasible iterations ensure convergence from remote initial guesses through a balance between optimality and feasibility.

Finally, we would like to emphasize that the feasibility-driven mode of \textsc{Box-\gls{fddp}} can not only solve tasks that \textsc{Box-\gls{ddp}}$^+$ is unable to solve, but also \textit{improve} the solutions of tasks that \textsc{Box-\gls{ddp}}$^+$ is able to solve.
For instance, consider the quadcopter tasks \textsc{goal}, \textsc{narrow}, and \textsc{loop}: In the accompanying video, we show that our approach generates concise and smooth quadcopter trajectories, whereas \textsc{Box-\gls{ddp}}$^+$ generates jerky motions and with unnecessary loops, due to early projection of the control commands.

\begin{table}%
\caption{Number of iterations, total cost, and average total computation time over 100 trials.}\vspace{-1em}
\label{tab:results}
\begin{center}
    \begin{tabular}{@{} l rcc r rcc @{}}
        \toprule
        &\multicolumn{3}{c}{\textsc{Box-\gls{ddp}}$^+$} && \multicolumn{3}{c}{\textsc{Box-\gls{fddp}} (feas)} \\
        \cmidrule{2-4}\cmidrule{6-8}
        \emph{Problems}  &Iter.     &Cost               &Time (s.) &   &Iter.    &Cost               &Time (s.)\\
        \midrule
         \textsc{pend}   &$105$     &$4.4292$           &$0.2473$  &   &$ 34$    &$0.4997$           &$0.0721$\\
         \textsc{goal}   &$ 23$     &$0.0764$           &$0.0913$  &   &$ 18$    &$0.0072$           &$0.0775$\\
         \textsc{loop} 	 &$133$     &$6.7211$           &$0.9144$  &   &$ 56$    &$0.6444$           &$0.3982$\\
         \textsc{narrow} &$ 70$     &$1.9492$           &$0.6781$  &   &$ 35$    &$0.4577$           &$0.3136$\\
         \textsc{man}    &$ 71$     &$4.6193$           &$11.428$  &   &$ 66$    &$4.6193$           &$7.7417$\\
         \textsc{taichi} &$120$     &$6.8184$           &$38.482$  &   &$101$    &$6.8184$           &$21.909$\\
         \textsc{jump}	 &$108$     &$1.34\textrm{e}5$  &\xmark    &   &$53$     &$6.67\textrm{e}4$  &$0.8226$\\
         \textsc{hop}    &$ 17$     &$1.3\textrm{e}12$  &\xmark    &   &$205$    &$1.91\textrm{e}4$  &$19.844$\\
         \textsc{dip}    &$121$     &$34.2$             &$106.9$   &   &$97$     &$34.2$             &$107.8$\\
         \textsc{pullup} &$176$     &$276.87$           &\xmark    &   &$426$    &$146.29$           &$422.5$\\
        \bottomrule
    \end{tabular}
\end{center}
\xmark\, algorithm does not find a solution.
\end{table}

\subsubsection{Gap contraction and nonlinearities}\label{sec:gap_analysis}
\begin{figure}[b]
  \centering
  \includegraphics[width=\linewidth]{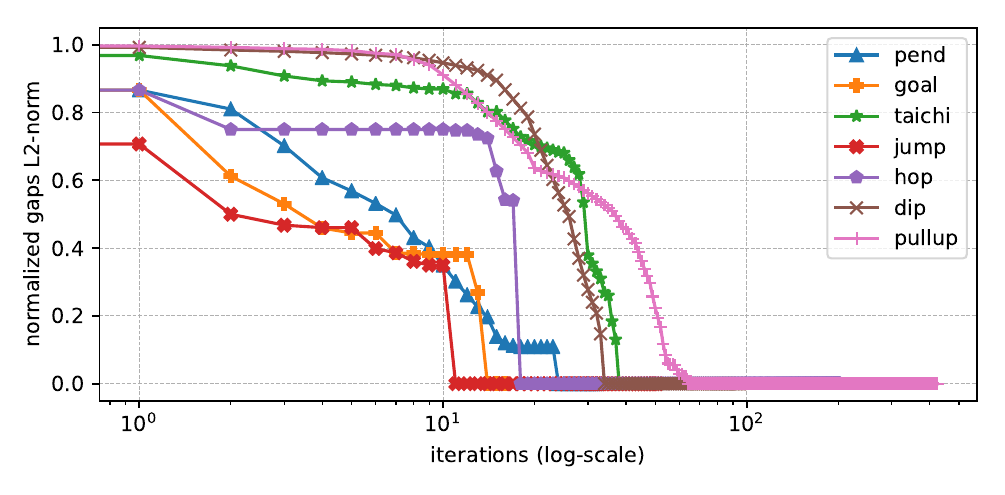}\vspace{-0.8em}
  \caption{Gap contraction of \textsc{Box-\gls{fddp}} for different optimal control problems.
  For all the cases, the gaps were open for the first several iterations.
  The gap contraction rate varies according to the accepted step-length.
  Smaller contraction rates, during the first iterations, appeared in very nonlinear problems (\textsc{taichi}, \textsc{man}, \textsc{hop}, and \textsc{jump}), because of the larger error of the search direction.}
  \label{fig:gap_contraction}
\end{figure}

We observed that the gap contraction rate is highly influenced by the nonlinearities of the system dynamics (see~\fref{fig:gap_contraction}).
When compared to the dynamics, the nonlinearities of the task often have a smaller effect (e.g., \textsc{dip} vs \textsc{pullup}).
Indeed, the gap contraction speed followed the order: humanoid, quadruped, double pendulum, and quadcopter.

Propagation errors due to the dynamics linearization have an important effect on the algorithm progress as Riccati recursions maintain a local quadratic approximation of the value function.
The prediction of the expected improvement is indeed more accurate for systems with less nonlinearities, which is the reason why the algorithm tends to accept larger steps that result in higher gap contractions.
Indeed, the effect of having a feasibility search is more significant in problems with very nonlinear dynamics as it reduces the total cost faster due to the nonlinearity distribution motivated in~\sref{sec:advantages_multshooting} (see~\fref{fig:cost_iterations} and~\ref{fig:gap_contraction}).
It also produces a low cost reduction rate during the gap contraction phase as the algorithm is first focusing on achieving dynamic feasibility.

\subsubsection{Highly-dynamic and complex maneuvers}\label{sec:highly_dynamic_maneuvers}
\begin{figure}%
  \centering
  \includegraphics[width=\linewidth]{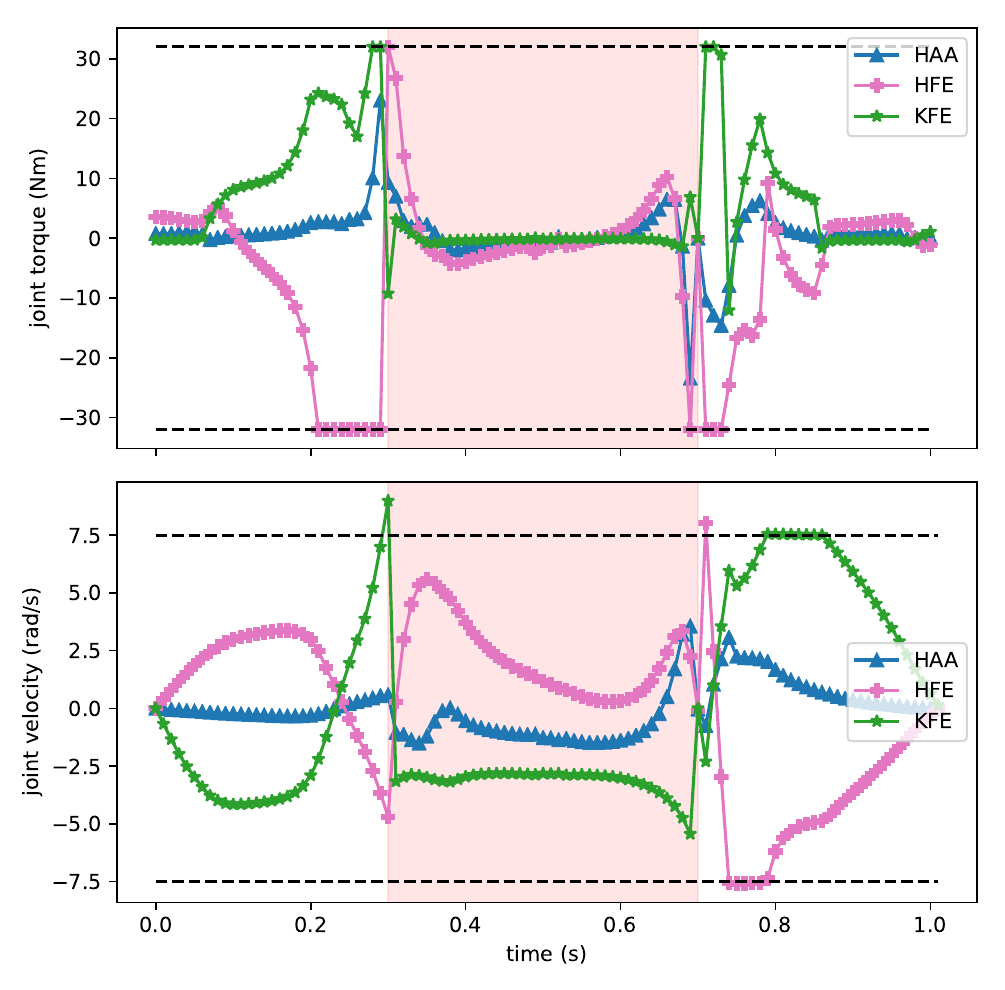}\vspace{-0.8em}
  \caption{Joint torques and velocities for the ANYmal jumping maneuver.
  (top) Generated torques of the \gls{lf} joints and its limits (32~\si{\newton\metre});
  (bottom) Generated velocities of the \gls{lf} joint and its limits (7.5~\si{\radian}/\si{\second}).
  The red region describes the flight phase.
  Note that HAA, HFE, and KFE are the abduction/adduction, hip flexion/extension and knee flexion/extension joints, respectively.}
  \label{fig:joint_torque_velocity}
\end{figure}

The \textsc{Box-\gls{fddp}} algorithm can solve a wide range of motions: from unstable and consecutive hops to aggressive and complex motor behaviors.
In~\fref{fig:joint_torque_velocity}, we show the joint torques and velocities of a single leg for the ANYmal's jumping problem (depicted in~\fref{fig:highly_dyn_maneuvers}c).
The motion consisted of three phases: jumping ($0$--$300$~\si{\milli\second}), flying ($300$--$700$~\si{\milli\second}), and landing ($700$--$1000$~\si{\milli\second}).
We used $0.7$ as a friction coefficient and reduced the real joint limits of the ANYmal robot: from $40$ to $32$~\si{\newton\metre} (torque limits) and from $15$ to $7.5$~\si{\radian}/\si{\second} (velocity limits).
Thus, generating a $30$~\si{\centi\metre} jump becomes a very challenging task.
Furthermore, in this experiment, the velocity limit violations appeared since we used quadratic penalization (with constant weights) to enforce them, and the swing phases are likely too short for such a large jump.
Note that if we use constant weights, it might turn out that, for some cases, these weights are not big enough.
Nonetheless, we only encountered these violations in very constrained problems.
For instance, we did not find velocity violations for the walking, trotting, pacing, and bounding gaits (reported in the accompanying video).
For these cases, \textsc{Box-\gls{fddp}} converged approximately with the same number of iterations achieved by the \gls{fddp} solver (i.e., a fully unconstrained case).
This is due to the fact that the robot could generate those gaits without reaching its torque limits.

Surprisingly, naturally looking behaviors emerged during the computation of the \textsc{dip} and \textsc{pullup} problems on the Talos humanoid robot.
We did not include any heuristic that could have helped the algorithm to generate these undefined behaviors.
For instance, the balancing and leg-crossing on the bars emerge if we allocate a significant amount of time in that motion phase.
Similarly, the pull-up motion emerges if we significantly increase the maximum torque limits on the arms.

\subsection{Experimental trials}
\begin{figure*}
  \centering
  \href{https://youtu.be/bOGBPTh_lsU?t=94}{\includegraphics[width=.95\linewidth]{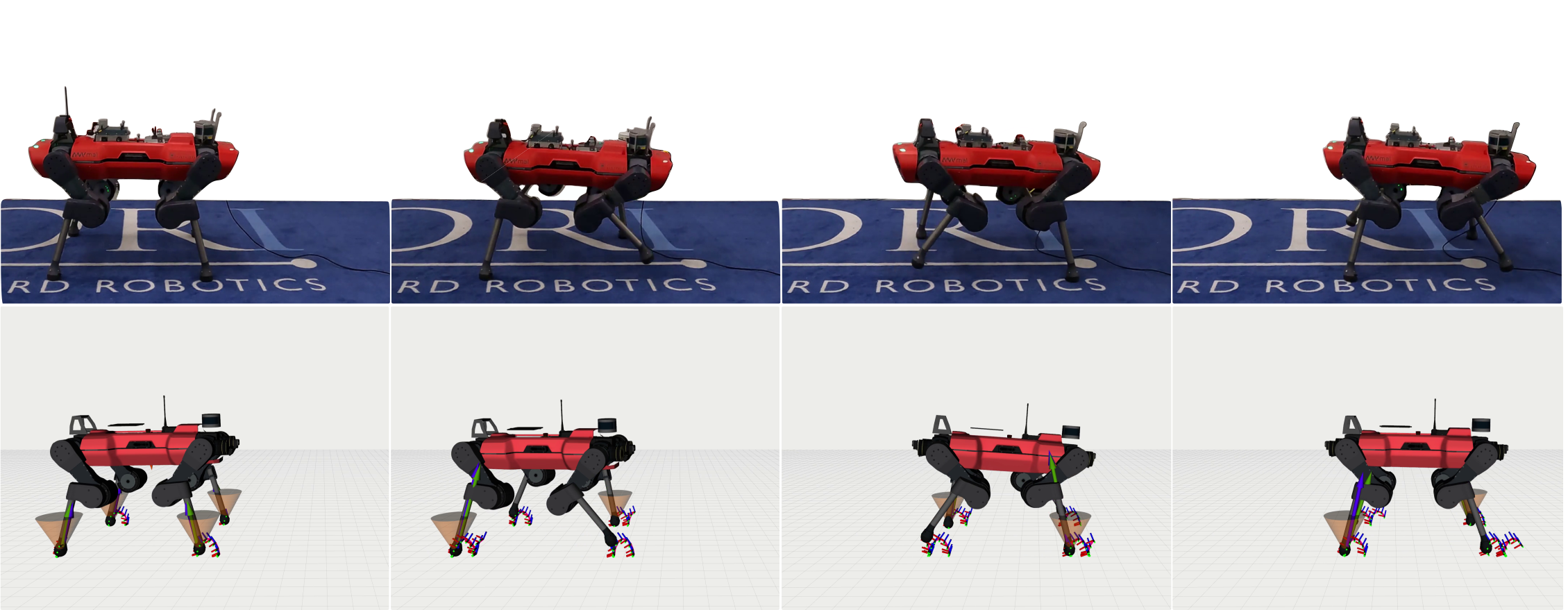}}
  \caption{Snapshots of hardware experiments using \textsc{Box-\gls{fddp}} on the ANYmal C quadruped: Here, the robot completes a trotting gait in a model predictive control fashion.
  Our method is able to resolve the optimal control problem for a \SI{1.25}{\second} horizon in \SI{\approx10}{\milli\second} on laptop hardware.
  The top figure shows snapshots of the ANYmal robot trotting forward, while the bottom figure displays the measured and desired contact forces and reference swing trajectories sent to the controller.
  To watch the video, click on the figure or see \texttt{\url{https://youtu.be/bOGBPTh_lsU}}.
  For more details about the predictive control algorithm, please refer to \cite{mastalli-tro22}.}
  \label{fig:hardware_experiments}
\end{figure*}

We demonstrated the capabilities of our \textsc{Box-\gls{fddp}} algorithm in a model predictive control application for the ANYmal C quadruped robot.
The \textsc{Box-\gls{fddp}} algorithm computed reference motions in real-time and its efficiency enables our predictive controller to run in excess of \SI{100}{\hertz} computation frequency with a horizon of \SI{1.25}{\second}.
\fref{fig:hardware_experiments} shows snapshots of the forward trotting gait computed with the \textsc{Box-\gls{fddp}} algorithm.
It also displays the contact-force tracking and updates of the swing-foot reference trajectories.
In this experiment, we predefined the timings for the swing-feet motions and the foothold locations.

\subsection{Box-FDDP vs a direct transcription formulation}\label{sec:knitro_comparison}
We formulated an efficient direct transcription problem with dynamics defects constraints in each node.
The formulation is conceptually similar to our feasible descent direction introduced in~\sref{sec:formal_guarantees}.
We transcribed the dynamics using a symplectic Euler scheme, the same integration scheme also used in~\textsc{Box-\gls{fddp}}.
We solved the direct transcription problem using different optimization algorithms provided by \textsc{Knitro}~\cite{byrd-knitro06}.
These available algorithms are: Interior/Direct~\cite{waltz-knitro06}, Interior/CG~\cite{byrd-knitro99}, active set~\cite{byrd-Knitro04} and \gls{sqp} algorithms.
Below we briefly describe each of \textsc{Knitro}'s algorithms, and then report the comparison results with \textsc{Box-\gls{fddp}}.

The Interior/Direct (\textsc{Knitro-IDIR}) algorithm replaces the \gls{nlp} problem with a series of barrier sub-problems.
In each iteration, it solves the primal-dual \gls{kkt} problem using a line search procedure.\footnote{For more details about interior point methods, the authors suggest the reader to see~\cite{nocedal-optbook}.}
Instead, Interior/CG (\textsc{Knitro-ICG}) solves the primal-dual \gls{kkt} problem using a projected conjugate gradient method.
This method uses exact second derivatives, without explicitly storing the Hessian matrix, through a tailored trust region procedure.
Interior/Direct also invokes this trust region procedure if the line search iteration converges to a non-stationary point~\cite{waltz-knitro06}.
In contrast to the interior point methods, the active set algorithm (\textsc{Knitro-SLQP}) replaces the \gls{nlp} problem with a sequence of quadratic programs to form a sequential linear-quadratic programming algorithm.
This algorithm selects a set of active constraints in each iteration, and produces a more exterior path (i.e., along the constraints) to the solution.
Finally, \textsc{Knitro}'s \gls{sqp} algorithm (\textsc{Knitro-SQP}) is also an active set method designed for small to medium scale problems with expensive function evaluations.
Both active set approaches are often preferable to interior point methods on small- to medium-sized problems when we can provide a good initial guess~\cite{nocedal-optbook}.
However, the problems in robotics are often large with many inequality constraints.
Indeed, the benefits of interior point methods have been pointed out in the context of direct methods~\cite{pardo-ral16}.

\subsubsection{Optimality vs feasibility}
We compared the total cost, number of iterations, and total computation time against the different algorithms implemented in \textsc{Knitro} over 100 trials.
For the comparison, we solved the double pendulum problem (\textsc{pend}), as it requires discovery of a swing-up maneuver.
With this, we can clearly compare the trade-off between optimality and feasibility across the different algorithms.
Note that, as described earlier, we used a single-thread for both \textsc{Knitro} and \textsc{Box-\gls{fddp}} despite our algorithm supporting multithreading.

\tref{tab:optimality_against_Knitro} reports three different formulations used in the \textsc{Knitro} algorithms.
The first one (\textsc{pen}) emulates exactly the optimal control formulation used in \textsc{Box-\gls{fddp}}, i.e., control constraints, regularization terms, and a terminal quadratic cost.
The second case (\textsc{regconst}) uses a terminal constraint to impose the desired up-ward position together with the regularization terms.
The third case (\textsc{const}) uses only the terminal constraints.
Below we summarize the obtained results for each formulation.

\textsc{Box-\gls{fddp}} converges faster (w.r.t. time) than \textsc{Knitro} algorithms in all of the above formulations.
However, \textsc{Knitro-ICG} is as fast as our approach with the \textsc{const} formulation.
On the other hand, when it comes to optimality, \textsc{Knitro} produces more optimal solutions if we use the \textsc{regconst} formulation.
Indeed, in our experience, \textsc{Knitro} generally has a better behavior when the formulation is dominated by constraint functions.
Note that we do not report the cost values for the \textsc{const} formulation as this boils down to a feasibility problem, i.e., a problem with only constraints.

\begin{table}[t]
  \caption{box-fddp vs Knitro in double pendulum problem.}\vspace{-1em}
  \label{tab:optimality_against_Knitro}
  \begin{center}
      \begin{tabular}{@{} llccc @{}}
          \toprule
          \emph{Case}           & \emph{Algorithms}           &Cost                &Iteration          &Total Time (sec.)  \\
          \bottomrule[0.5pt]\\[-5pt]
                                &\textsc{boxfddp}             &$0.4997$            &$34$               &$\textbf{0.0721}$ \\
           \midrule[0.1pt]
           \textsc{pen}         &\textsc{Knitro-IDIR}         &$2.1094$            &$1064$             &$8.6704$  \\
           \textsc{regconst}    &\textsc{Knitro-IDIR}         &$\textbf{0.4693}$   &$47$               &$0.2766$  \\
           \textsc{const}       &\textsc{Knitro-IDIR}         &--                  &$20$               &$0.1387$  \\
           \midrule[0.1pt]
           \textsc{pen}         &\textsc{Knitro-ICG}          &$0.5441$            &$59$               &$0.2390$  \\
           \textsc{regconst}    &\textsc{Knitro-ICG}          &$0.4787$            &$50$               &$0.1875$  \\
           \textsc{const}       &\textsc{Knitro-ICG}          &--                  &$\textbf{16}$      &$0.0733$  \\
           \midrule[0.1pt]
           \textsc{pen}         &\textsc{Knitro-SLQP}         &$0.5978$            &$165$              &$1.2595$  \\
           \textsc{regconst}    &\textsc{Knitro-SLQP}         &$0.4773$            &$287$              &$1.6704$  \\
           \textsc{const}       &\textsc{Knitro-SLQP}         &--                  &$27$               &$0.2359$  \\
           \midrule[0.1pt]
           \textsc{pen}         &\textsc{Knitro-SQP}          &$0.5889$            &$114$              &$46.88$  \\
           \textsc{regconst}    &\textsc{Knitro-SQP}          &$0.4767$            &$182$              &$12.233$  \\
           \textsc{const}       &\textsc{Knitro-SQP}          &--                  &$22$               &$8.7736$  \\
          \toprule
      \end{tabular}
  \end{center}
\end{table}

\begin{figure}[t]
  \centering
  \includegraphics[width=\linewidth]{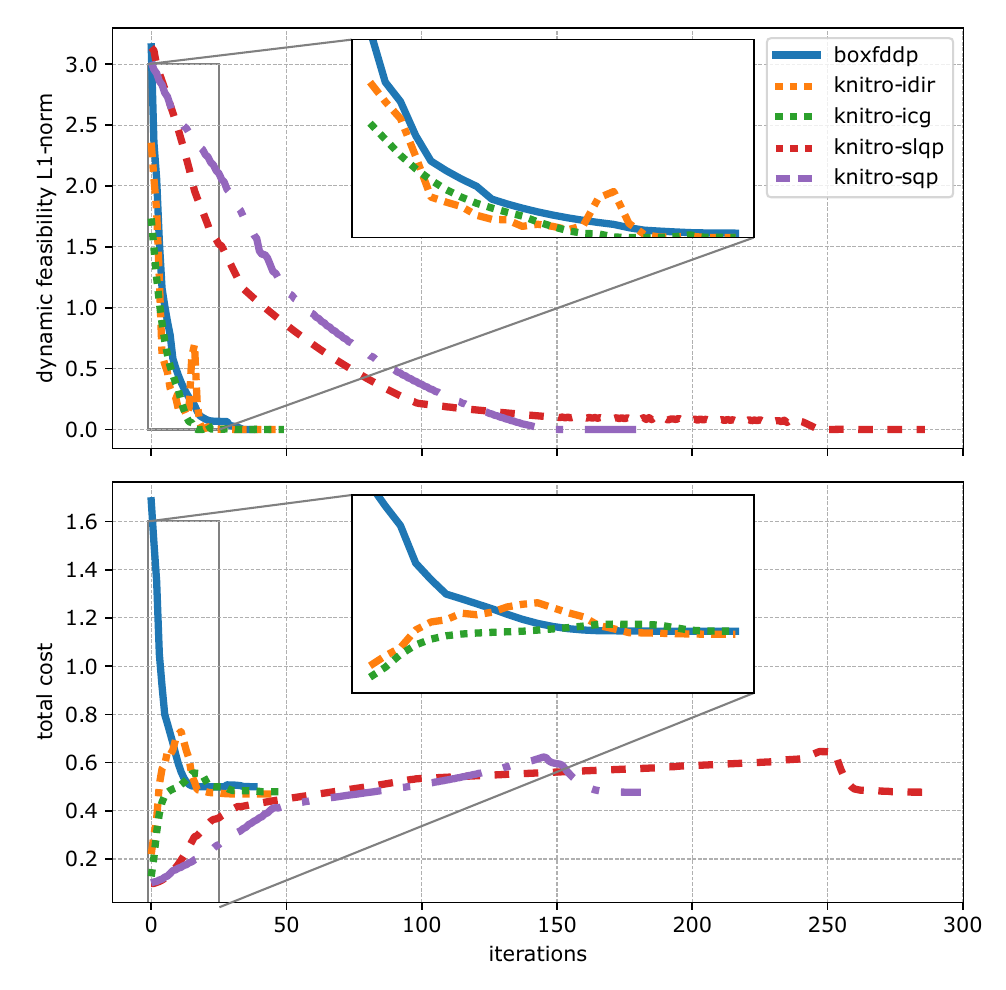}\vspace{-0.8em}
  \caption{Dynamic feasibility and cost evolution for the double pendulum problem.
  \textsc{Box-\gls{fddp}} had a similar evolution (dynamic feasibility and cost) to the interior point algorithms implemented in \textsc{Knitro}. 
  \textsc{Knitro} produced lower cost solutions, but the computational burden is often much higher.}
  \label{fig:feasibility_optimality_comparison}
\end{figure}

We used the \textsc{regconst} formulation to be able to compare both: cost and feasibility evolution.
The dynamic infeasibility decreased monotonically for all the algorithms as plotted in~\fref{fig:feasibility_optimality_comparison} (top).
\textsc{Box-\gls{fddp}} shows a fast resolution of the dynamics feasibility as interior point algorithms that, generally speaking, require fewer iterations than active set approaches~\cite{hei-Knitro08}.
The cost evolution is also similar to the interior point algorithms, where the total costs are reported as \textsc{regconst} cases in~\tref{tab:optimality_against_Knitro}.

\subsubsection{Computation time}
\textsc{Box-\gls{fddp}} had a better runtime performance than the \textsc{Knitro} algorithms for the double pendulum problem (cf.~\tref{tab:optimality_against_Knitro}).
However, to answer the runtime performance scalability to higher-dimensional optimal control problems, we analyzed the problem of generating a forward jumping maneuver with the ANYmal robot (i.e., \textsc{jump}).

\begin{table}[t]
  \caption{Runtime performance for a forward jumping maneuver over 100 trials.}\vspace{-1em}
  \label{tab:computation_time_against_Knitro}
  \begin{center}
      \begin{tabular}{@{} c cc r cc @{}}
          \toprule
          &\multicolumn{2}{c}{\textsc{Knitro-IDIR}} && \multicolumn{2}{c}{\textsc{Box-\gls{fddp}}} \\
          \cmidrule{2-3}\cmidrule{5-6}
          \emph{Trunk height (m.)}  &Iter.     &Time (sec.)  &   &Iter.      &Time (sec.)\\
          \midrule
          $0.25$                     &\textbf{11}        &1.5039      &   &59         &\textbf{0.9061}\\
          $0.30$                     &\textbf{10}        &1.3900      &   &72         &\textbf{1.1409}\\
          $0.35$                     &\textbf{15}        &1.9565      &   &59         &\textbf{0.9030}\\
          $0.40$                     &\textbf{18}        &2.3864      &   &49         &\textbf{0.7849}\\
          $0.45$                     &\textbf{16}        &2.1320      &   &51         &\textbf{0.8574}\\
          \bottomrule
      \end{tabular}
  \end{center}
\end{table}

We used the same phase timings, joint limits and friction coefficient reported in~\sref{sec:highly_dynamic_maneuvers}.
The results reported with \textsc{Knitro} and \textsc{Box-\gls{fddp}} cases are based on slightly different optimal control formulations.
The idea is to define the most suitable formulation for each algorithm.
For instance, we use quadratic penalization terms to impose the desired foothold placement, joint velocity limits and friction cone constraints for the \textsc{Box-\gls{fddp}} algorithm.
Instead, for the \textsc{Knitro} algorithms, we substitute these penalization terms by general equality and inequality constraints.
To further reduce the computation time of \textsc{Knitro} cases, we also impose a constraint for the terminal position of the trunk.
Note that we did not include any cost term since it negatively affects the convergence rate of \textsc{Knitro}, i.e., we treated it as a feasibility problem.

\tref{tab:computation_time_against_Knitro} reports the runtime performance over $100$ trials for the \textsc{Knitro-IDIR} algorithm only.
The other methods (i.e., \textsc{Knitro-ICG}, \textsc{Knitro-SLQP} and \textsc{Knitro-SQP}) were unable to solve this problem.
We used 5 different initial trunk heights, and we initialized the algorithms using their corresponding joint posture (as described in Appendix~\ref{sec:quadruped_problems}) and no controls (i.e., $\mathbf{u}^0_s=\{\mathbf{0},\cdots,\mathbf{0}\}$).
As in the double pendulum case, \textsc{Box-\gls{fddp}} also solved this problem faster than \textsc{Knitro} algorithms, even though it required a significant number of extra---computationally inexpensive---iterations.
We suspect that this increment in the number of iterations is due to the use of penalization terms in the contact placement, friction cone and state limits constraints.

\subsubsection{Robustness against different initial guesses}
We compared the robustness against different initial guesses for the double pendulum (\textsc{pend}) and quadrupedal jump (\textsc{jump}) problems.
In both problems, we generated random joint postures---around the nominal state---and used them to define an initial guess for the state trajectory $\mathbf{x}^0_\mathbf{s}$.
In addition to the robot's joint postures, we also generated random joint velocities around the \textit{zero-velocity} condition for the double pendulum case only.
We used this single random posture and velocity for each node in $\mathbf{x}^0_\mathbf{s}$, and initialized the control sequence with zeros.
We used the most suitable formulations for \textsc{Box-\gls{fddp}} and \textsc{Knitro} algorithms as justified above.

\tref{tab:robustness_warm_starting} reports the number of successful resolutions over $100$ trials.
We considered a problem to be successfully solved if the gradient of the \textsc{Box-\gls{fddp}} or the feasibility of the \textsc{Knitro} algorithms are lower than $5\times 10^{-5}$ (absolute feasibility tolerance).
Note that this includes the cases where \textsc{Knitro} found a feasible approximate solution.\footnote{For further detail, we suggest the reader to consult the \textsc{Knitro} manual:~\url{https://www.artelys.com/docs/knitro/3_referenceManual.html}.}
Furthermore, we considered a problem resolution unsuccessful if the problem does not converge within \SI{70}{\second}, which is enough time as we can see above.
For each problem, we used two different maximum values of the random initialization, which their maximum magnitude are described using the $\ell^\infty$ norm (i.e., $\|\cdot\|_\infty$).
We added this additive noise to the default initial guess (described in Appendix~\ref{sec:oc_problems}) used for the state trajectory.

As expected, the \textsc{Knitro} interior point methods performed better than the active set ones.
For the double pendulum problem, the interior point algorithms (i.e., \textsc{Knitro-IDIR} and \textsc{Knitro-ICG}) perform better than \textsc{Box-\gls{fddp}} if the warm-starting point is close to the initial condition.
Despite that, \textsc{Box-\gls{fddp}} shows more robustness to initial guesses as its percentage of successful resolutions is consistent.
Furthermore, we observed a significant increment in the number of successful resolutions for the \textsc{jump} problem.
Indeed, \textsc{Knitro} was not able to solve this problem at all for random magnitudes bigger than $\|0.01\|_\infty$.

\begin{table}[b]
  \caption{Percentage of successful resolutions from random initial guesses.}\vspace{-1em}
  \label{tab:robustness_warm_starting}
  \begin{center}
      \begin{tabular}{@{} l cc c cc @{}}
          \toprule
          &\multicolumn{2}{c}{\textsc{pend}} && \multicolumn{2}{c}{\textsc{jump}} \\
          \cmidrule{2-3}\cmidrule{5-6}
          \emph{Algorithms}           &$\|1\|_\infty$     &$\|100\|_\infty$     &     &$\|0.0005\|_\infty$     &$\|0.005\|_\infty$\\
          \midrule
               \textsc{Box-FDDP}       &$74\,\%$           &$\textbf{44}\,\%$    &     &$\textbf{99}\,\%$       &$\mathbf{99}\,\%$\\ 
               \textsc{Knitro-IDIR}	  &$\textbf{100}\,\%$ &$1\,\%$              &   &$51\,\%$                  &$13\,\%$\\
               \textsc{Knitro-ICG}	  &$\textbf{100}\,\%$ &$0\,\%$              &   &\xmark                    &\xmark\\
               \textsc{Knitro-SLQP}	  &$95\,\%$           &$0\,\%$              &   &\xmark                    &\xmark\\
               \textsc{Knitro-SQP}	  &$92\,\%$           &$0\,\%$              &   &\xmark                    &\xmark\\
          \bottomrule
      \end{tabular}
  \end{center}
 \xmark\, algorithm does not find a solution.
\end{table}

\subsection{Box-FDDP, Box-DDP, and squashing approach in nonlinear problems}\label{sec:squashing_analisis}
To evaluate the numerical performance of having two modes, we compared the \textsc{Box-\gls{fddp}} (with two modes depending on the dynamics feasibility), \textsc{Box-\gls{ddp}}$^+$ (using a single mode) and \gls{ddp}$^+$ with a squashing function (using a single mode) for three scenarios with the IRIS quadcopter: reaching goal (\textsc{goal}), looping maneuver (\textsc{loop}), and traversing a narrow passage (\textsc{narrow}).
We used a sigmoidal element-wise squashing function of the form:
\begin{eqnarray*}
  \mathbf{s}^i(\mathbf{u}^i) = \cfrac{1}{2}\left(\mathbf{\underline{u}}^i + \sqrt{\gamma^2 + (\mathbf{u}^i-\mathbf{\underline{u}}^i)^2}\right) +\\
   \cfrac{1}{2}\left(\mathbf{\overline{u}}^i - \sqrt{\gamma^2 + (\mathbf{u}^i-\mathbf{\overline{u}}^i)^2}\right)
\end{eqnarray*}
in which the sigmoid is approximated through two smooth-abs functions, $\gamma$ defines its smoothness, and $\mathbf{\underline{u}}^i$, $\mathbf{\overline{u}}^i$ are the element-wise lower and upper control bounds, respectively.
We introduced this squashing function on the system controls as: $\mathbf{x}_{k+1} = \mathbf{f}(\mathbf{x}_k,\mathbf{s}(\mathbf{u}_k))$.
We used $\gamma=2$ for all the experiments presented in this section.

\begin{figure}[t]
  \centering
  \includegraphics[width=\linewidth]{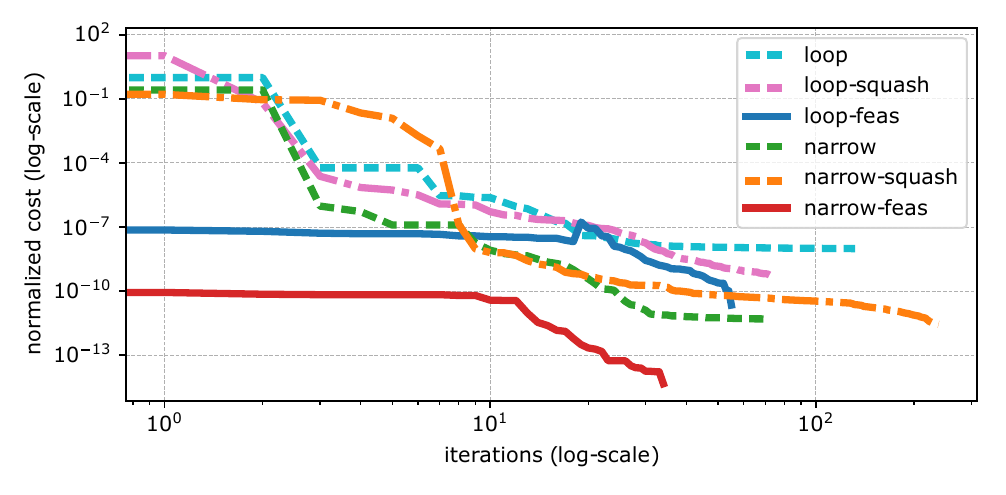}\vspace{-0.8em}
  \caption{Cost and convergence comparison for different quadcopter maneuvers: looping (\textsc{loop}) and narrow passage traversing (\textsc{narrow}).
  \textsc{Box-\gls{fddp}} (*-feas) outperformed both \textsc{Box-\gls{ddp}}$^+$ and \gls{ddp} with squashing function (*-squash).}
  \label{fig:cost_iterations_squash}
\end{figure}

\fref{fig:cost_iterations_squash} shows that \textsc{Box-\gls{fddp}} converged faster than the other approaches.
As reported in the accompanying video, \textsc{Box-\gls{fddp}} did not generate undesired loops and jerky motions as in the other cases.
Indeed, the solutions with \textsc{Box-\gls{fddp}} have the lowest cost values (cf.~\tref{tab:results}).
We also observed that the squashing approach often converges sooner compared to \textsc{Box-\gls{ddp}}$^+$.
The main reason is due to the early saturation of the controls performed by \textsc{Box-\gls{ddp}}$^+$.

\begin{figure}[b]
  \centering
  \includegraphics[width=\linewidth]{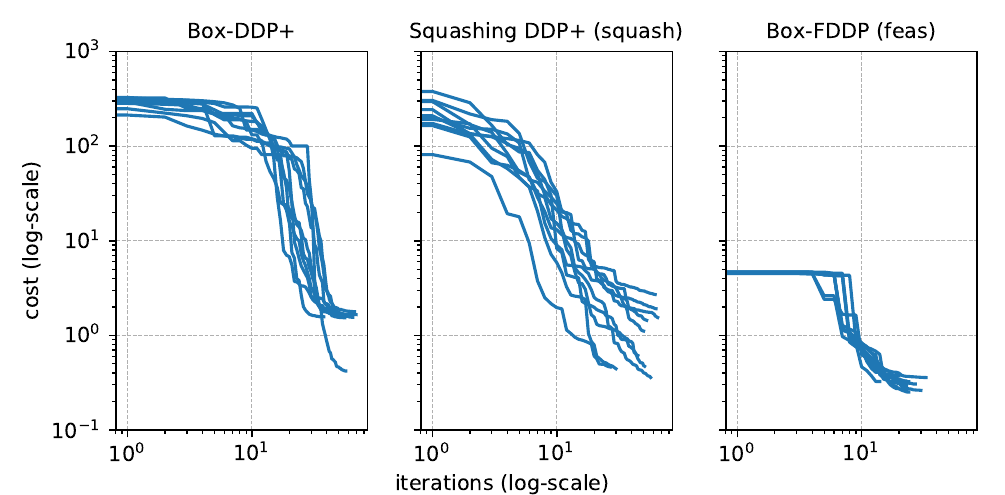}\vspace{-0.8em}
  \caption{Costs associated for 10 different initial configurations of reaching goal tasks.
  \textsc{Box-\gls{fddp}} converges earlier and with lower total cost than \textsc{Box-\gls{ddp}}$^+$ and \gls{ddp}$^+$ with squashing function.
  The performance of the squashing function approach exhibits a high dependency on the initial condition.}
  \label{fig:cost_iterations_initial_conditions}
\end{figure}

In \fref{fig:cost_iterations_initial_conditions}, we show the cost evolution for $10$ different initial configurations of the reaching goal task.
The target and initial configurations are $(3, 0, 1)$ and $(-0.3 \pm 0.6, 0, 0)$ \si{\metre}, respectively.
Infeasible iterations, in \textsc{Box-\gls{fddp}}, produce a very low cost in the first iterations.
The squashing approach is the most sensitive to initial conditions.
However, on average, it produces slightly better solutions than \textsc{Box-\gls{ddp}}$^+$.
This is in contrast to the reported results in~\cite{tassa-icra14}, where the performance was analyzed only for the linear-quadratic regulator problem.

\section{Conclusion}\label{sec:conclusion}
We proposed a feasibility-driven approach whose search is primarily driven by the dynamics feasibility of the optimal control problem.
The dynamically-infeasible iterations, which mimic a direct multiple shooting approach, allowed us to solve a wide range of optimal control problems despite it being provided with a poor initialization.
The benefits of our approach are crystallized over a set of athletic and highly-dynamic maneuvers computed for the Talos humanoid and the ANYmal quadruped robots, respectively.
Its improvement on the basin of attraction for a good local optimum has been a key factor to optimize such kind of complex maneuvers while considering the robot's full rigid body dynamics, joint limits and friction cone constraints.
Indeed, \textsc{Box-\acrshort{fddp}} has shown an increment in the robustness against different initial guesses compared with advanced \textsc{Knitro} algorithms.

We have provided evidence that our algorithm produces descent search directions.
For instance, we have observed that the feasibility contraction decreases monotonically as often happens in the advanced nonlinear programming algorithms available in \textsc{Knitro}.
Our approach has also shown to quickly reduce the dynamic infeasibility as observed in the most competitive \textsc{Knitro} algorithms.
A similar effect is observed in the cost evolution as well.
Our results suggest that the gap contraction rate is influenced by the nonlinearities of the system dynamics.

The runtime performance of \textsc{Box-\acrshort{fddp}} is often superior to direct transcription solved using state-of-the-art \textsc{Knitro} algorithms despite the increase in number of iterations due to the use of quadratic penalization terms.
However, when comparing the computation time per iteration, our approach is between $2$ to $10$ times faster, which makes it suitable for model predictive control applications.
Indeed, we demonstrated that our \textsc{Box-\gls{fddp}} algorithm can generate trotting gaits on the ANYmal C robot in a predictive control fashion.
One additional remark is that we have not considered the runtime reduction due to code generation support in \textsc{Crocoddyl} via \textsc{CppADCodeGen}~\cite{leal-cppadcodegen} and \textsc{CppAD}~\cite{bell-cppad12}.
According to our experience, code generation can lead to a computation time reduction between \SIrange{30}{60}{\percent} as can be seen in our public benchmarks~\cite{crocoddylweb}.

We have developed and reported the results for a wide range of optimal control problems in robotics.
These problems cover an important part of the spectrum of robotics applications.
The comparison across all these problems is unusual, and it represents an important contribution to the research community since we have open sourced many of these examples, as well as the~\textsc{Box-\gls{fddp}} algorithm, in the \textsc{Crocoddyl} repository~\cite{crocoddylweb}.
Our feasibility-driven approach enabled model predictive control applications on the ANYmal C robot, however, it can potentially be used in other applications such as in humanoid robotics~\cite{eber21ichr}, robot co-design~\cite{traiko22iros}, and learning~\cite{lembono20icra}.

\section*{Acknowledgments}
The authors are grateful to Matt Timmons-Brown and Vladimir Ivan, from the University of Edinburgh, for the production of the audio material of our video and the fruitful discussions on the study of the effects of our feasibility-driven approach, respectively.
\appendices
\section{Optimal control problems}\label{sec:oc_problems}
We divide the optimal control problems into four subsections: double pendulum, quadcopter, quadruped and humanoid robots.

\subsection{Double pendulum (pend)}
The goal is to swing from the stable to the unstable equilibrium points, i.e. from down-ward to up-ward positions, respectively.
To increase the problem complexity, the double pendulum (with weight of $\approx$~$4.5$~\si{\newton}) has a single actuated joint with small range of control (from $-5$ to~\SI{5}{\newton\metre}, largely insufficient for a static execution of the trajectory).
The time horizon is \SI{1}{\second} with $100$ nodes.
We define a quadratic cost function, for each node, that aims to reach the up-ward position.
For the running and terminal nodes, we use the weight values of $2\times 10^{-4}$ and $2\times 10^4$, respectively.
Additionally, we provide state and control regularization terms.
To inspect the algorithm capabilities, we \textit{do not provide} an initial guess (i.e., zeros for the states and controls), thus the swing-up strategy is discovered by the solver itself.
Finally, we implement the cost, dynamics, and their analytical derivatives.

\subsection{Quadcopter}
We consider three tasks for the IRIS quadcopter: reaching goal (\textsc{goal}), looping maneuver (\textsc{loop}), and traversing a narrow passage (\textsc{narrow}).
We define different way-points to describe the tasks, where each way-point specifies the desired pose and velocity.
The way-points are described through cost functions in the robot placement and velocity at specific instants of the motion.
These cost functions are quadratic term with $10^2$ as the weight value for both: pose and velocity.
We also include quadratic regularization terms for the state and control, their weight values are $10^{-5}$ and $10^{-1}$, respectively. 
The vehicle pose is described as a $\mathbb{SE}(3)$ element, which allows us to consider any kind of motion such as \textit{looping} maneuvers.
Control inputs are considered to be the thrust produced by the propellers, which can vary within a range from $0.1$ to~\SI{10.3}{\newton} each.
We compute the dynamics using the \gls{aba} algorithm\footnote{For more details about the \gls{aba} algorithm see~\cite{featherstone-rbdbook}.}, and the analytical derivatives are calculated as described in~\cite{carpentier-rss18}.
We integrate the dynamics with a fixed time step of $10$~\si{\milli\second}.
The solution is computed from a cold-start of the algorithm.
As in the double pendulum case, we do not provide an initial guess to the solvers.

\subsection{Aggressive jump, unstable hopping, and various gaits}\label{sec:quadruped_problems}
We use the ANYmal quadruped robot to generate a wide range of motions---jumping, hopping, walking, trotting, pacing, and bounding.
We deliberately reduce the torque and velocity limits to~\SI{32}{\newton\metre} and~\SI{7.5}{\radian}/\si{\second}, respectively, which intentionally increases the complexity of the jumping task (\textsc{jump}).
We use a quadratic barrier to penalize the joint velocities and the contact forces that are outside the limits.
The unstable hopping problem (\textsc{hop}) has a long horizon: \SI{7.14}{\second} with $714$ nodes.
It includes $10$ hops in total with a phase that switches the feet in contact.
Instead, the other problems have a horizon between $0.7-$\SI{1}{\second} with $70-100$ nodes.
We define quadratic terms, with a weight value of $10^6$, to track the desired swing-foot placements for each case.
We regularize the state trajectory around the robot's default configuration.
We use a friction coefficient of $0.7$.
Furthermore, we formulate a multi-phase optimal control problem using rigid contact dynamics, and their analytical derivatives, as described in~\cite{budhiraja-ichr18,mastalli-icra20}, respectively.
During a contact transition, we employ the impulse dynamics with analytical derivatives as also described in~\cite{mastalli-icra20}.
Our contact dynamics model also includes the Baumgarte gains, which are needed to numerically stabilize differential algebraic equations~\cite{baumgarte-72}; we use the values: $(0, 50)$.
We initialize the solver using the default posture and the quasi-static torques for each node of the initial guess trajectories.
The default posture defines the standing position of the robot; it does not provide any relevant information for a specific maneuver (e.g., \textsc{jump}).
The quasi-static torques describe the forces required to cancel the effects of gravity subject to the robot's default posture.

\subsection{Whole-body manipulation and balance}
We consider four problems for the Talos humanoid robot: whole-body manipulation (\textsc{man}), hand control while balancing in single leg (\textsc{taichi}), dip on parallel bars (\textsc{dip}) and a pull-up bar task (\textsc{pullup}).
For the dip and pull-up workouts, we define the real torque limits for the leg and torso joints.\footnote{Talos' arms are not strong enough to support its own weight.}
Additionally, we consider joint position and velocity limits in each scenario through a quadratic barrier.
Both \textsc{taichi}, \textsc{dip} and \textsc{pullup} tasks are divided into three phases; for the \textsc{taichi} task: manipulation, standing on one foot, and balancing; for the \textsc{dip} and \textsc{pullup} bar task: grasping the bar, workout, and landing on ground.
\textsc{Dip} and \textsc{pullup} problems have a horizon of \SI{12}{\second} with $400$ nodes.
The horizon of the whole-body manipulation and \textsc{taichi} problem are $3$ and \SI{6}{\second} with $60$ and $120$ nodes, respectively.
We define different Baumgarte gains for the hands and feet, i.e. $(0, 20)$ and $(0, 40)$, respectively.
We use the contact-placement cost functions for both: feet and hands.
Indeed, behaviors such as dip / pull-up balancing, pull-up motion, or the leg-crossing strategy emerge from our optimization algorithm since we do not describe the center of mass and swing-contact motions.
Finally, the regularization terms, dynamics and initialization strategy are the same that we use for the ANYmal cases.
For the details about the ANYmal problems see above~(\sref{sec:quadruped_problems}).

\bibliography{references}

\begin{IEEEbiography}[{\includegraphics[width=1in,height=1.25in,clip,keepaspectratio]{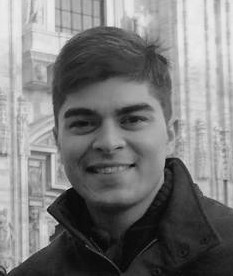}}]
	{Carlos Mastalli} received the M.Sc. degree in mechatronics engineering from the Simón Bolívar University, Caracas, Venezuela, in 2013 and the Ph.D. degree in bio-engineering and robotics from the Istituto Italiano di Tecnologia, Genoa, Italy, in 2017.

	He is currently an Assistant Professor at Heriot-Watt University, Edinburgh, U.K. He is the Head of the Robot Motor Intelligence (RoMI) Lab affiliated to the National Robotarium and Edinburgh Centre for Robotics. He is also appointed as Research Scientist at IHMC, USA. Previously, he conducted cutting-edge research in several world-leading labs: Istituto Italiano di Tecnologia (Italy), LAAS-CNRS (France), ETH Zürich (Switzerland), and the University of Edinburgh (UK).	His research focuses on building athletic intelligence for robots with legs and arms. To do so, he is working at the intersection of model predictive control, numerical optimization, deep learning, machine learning, whole-body control, and robot co-design.
\end{IEEEbiography}

\begin{IEEEbiography}[{\includegraphics[width=1in,height=1.25in,clip,keepaspectratio]{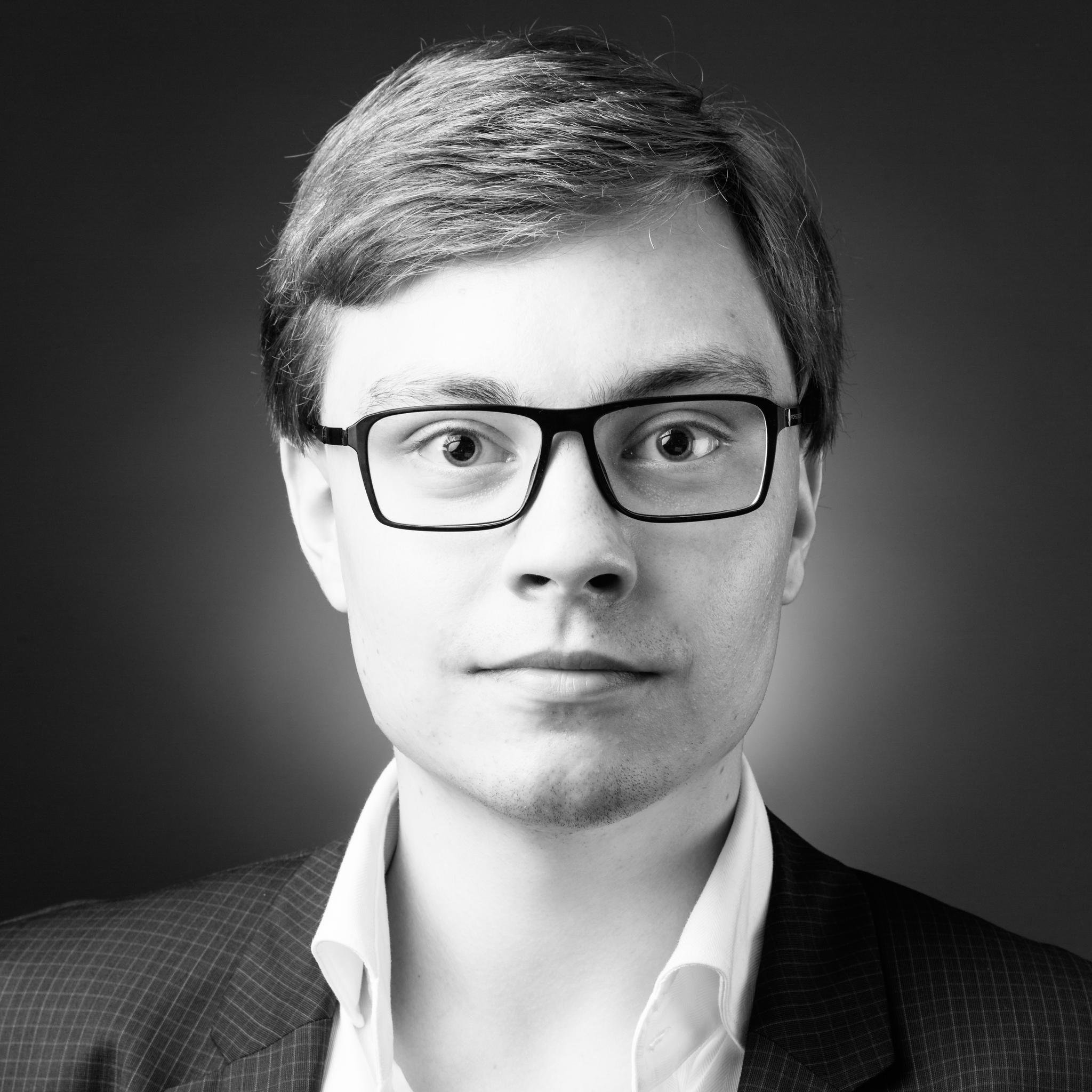}}]
	{Wolfgang Merkt} received the B.Eng.(Hns) degree in mechanical engineering with management and the M.Sc.(R) and Ph.D. degrees in robotics and autonomous systems from the University of Edinburgh, Edinburgh, U.K., in 2014, 2015 and 2019, respectively.

	He is currently a Postdoctoral Researcher at the Oxford Robotics Institute, University of Oxford with I. Havoutis.
	During his Ph.D., he worked on trajectory optimization and warm starting optimal control for high-dimensional systems and humanoid robots under the supervision of S. Vijayakumar.
	His research interests include fast optimization-based methods for planning and control, loco-manipulation, and legged robots.
\end{IEEEbiography}

\begin{IEEEbiography}[{\includegraphics[width=1in,height=1.25in,clip,keepaspectratio]{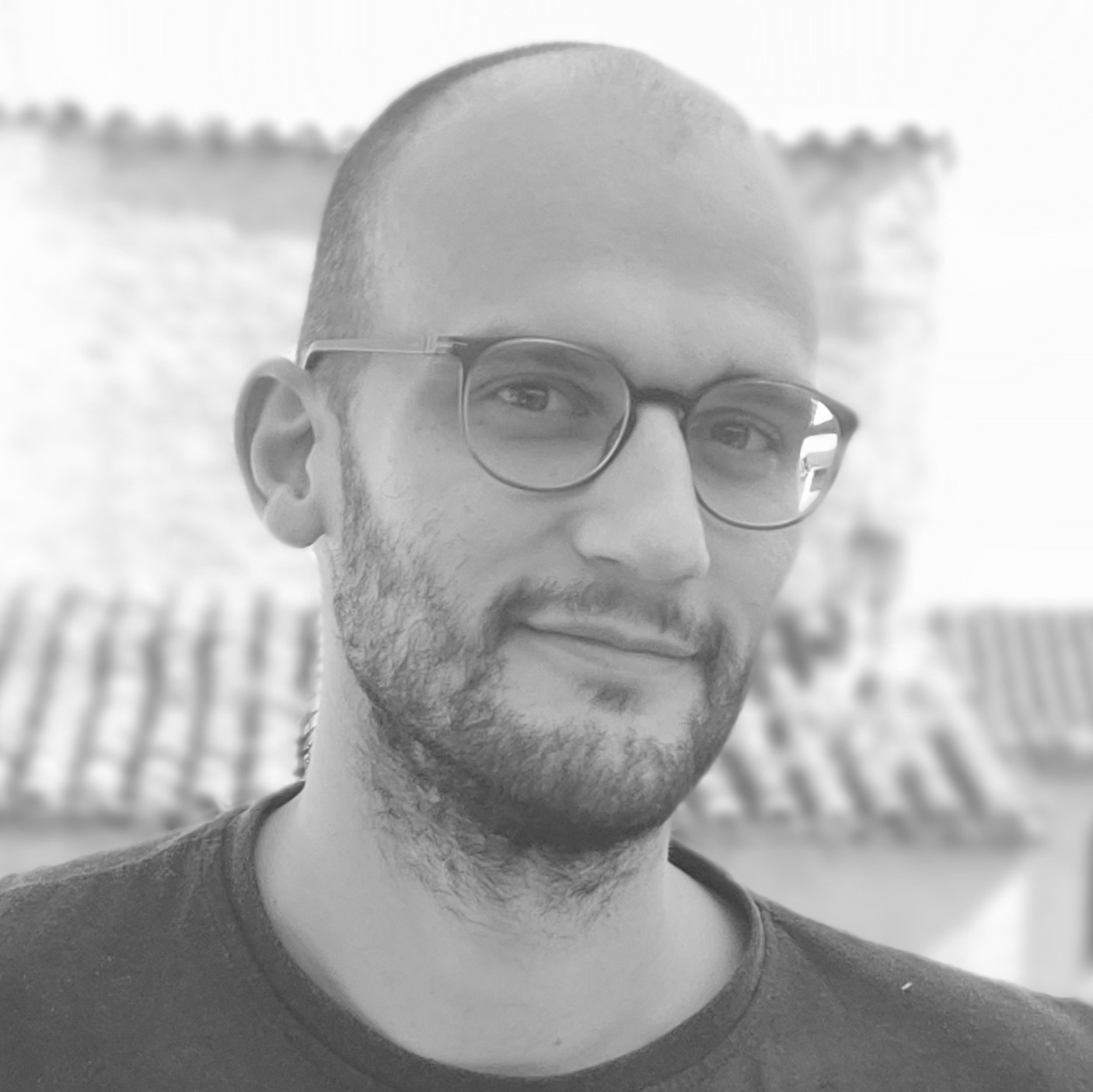}}]
	{Josep Marti-Saumell} received the B.Sc. degree in industrial engineering (majoring in mechanics) and the M.Sc. degree in automatic control and robotics from the Universitat Politècnica de Catalunya, Barcelona, Spain, in 2013 and 2018, respectively.

	He is currently a Ph.D. candidate at the Institut de Robòtica i Informàtica Industrial, CSIC-UPC, Barcelona, Spain.
	His current research interests include optimal control applied to mobile robotics, numerical optimization, and unmanned aerial manipulators.
\end{IEEEbiography}

\begin{IEEEbiography}[{\includegraphics[width=1in,height=1.25in,clip,keepaspectratio]{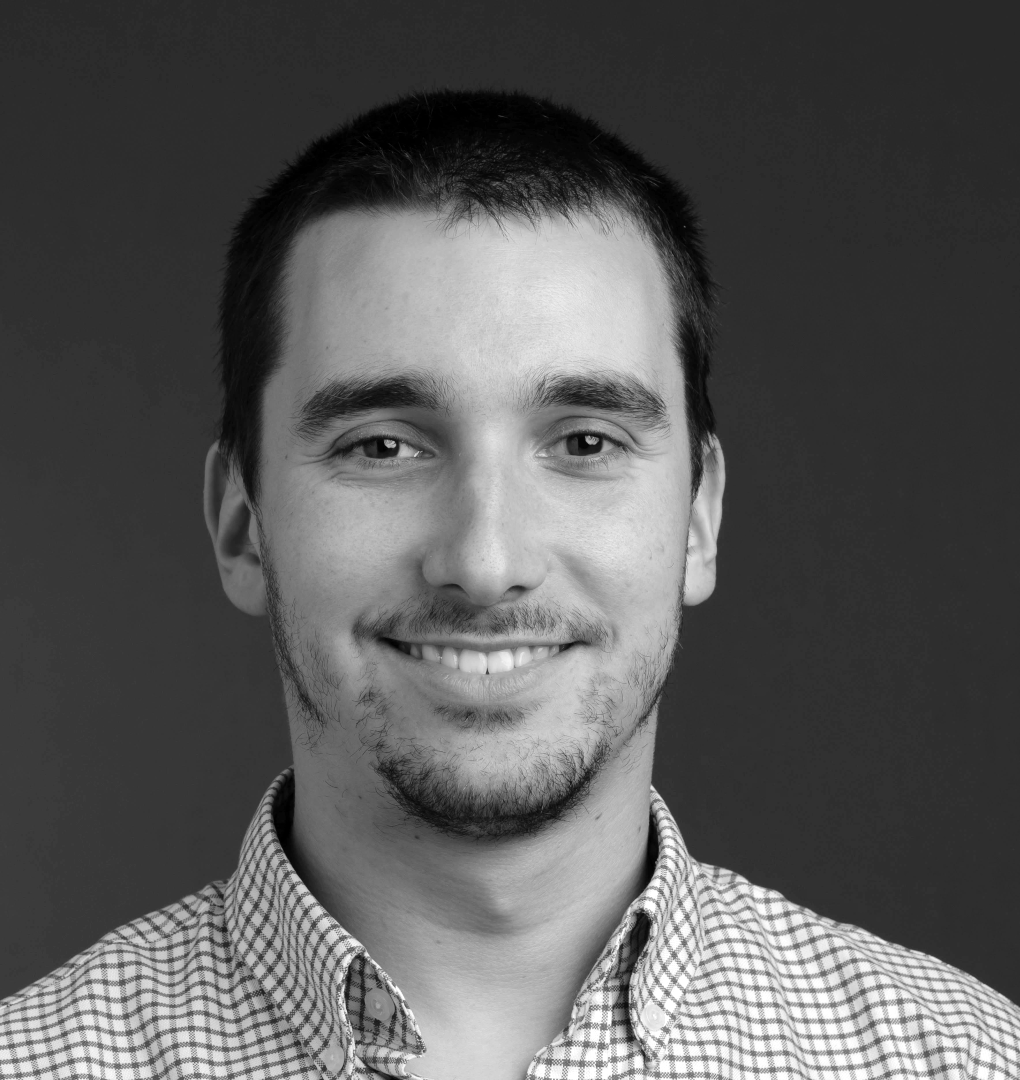}}]
	{Henrique Ferrolho} received his M.Sc. degree in informatics and computing engineering from the University of Porto, Porto, Portugal in 2017.

	He is currently pursuing a Ph.D. degree in robotics and autonomous systems at the University of Edinburgh under the supervision of S. Vijayakumar.
	His research interests include robust motion planning, and optimal control of agile robotic systems.
\end{IEEEbiography}
\begin{IEEEbiography}[{\includegraphics[width=1in,height=1.25in,clip,keepaspectratio]{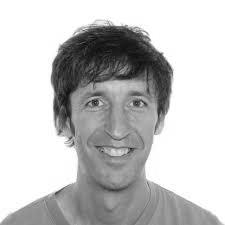}}]
	{Joan Solà} received the M.Sc. degree in telecommunication and electronics from the Universitat Politècnica de Catalunya, Barcelona, Spain, and the Ph.D. degree in robotics from the University of Toulouse, Toulouse, France, in 2007.

	He is currently a CSIC Researcher with the Institut de Robòtica i Informàtica Industrial, CSIC-UPC, Barcelona, Spain.
	He has also worked in the industry in the renewable energies sector, and was involved in the construction of a manned submarine  for depths up to 1200 m.
	He has contributed to monocular SLAM, especially in the undelayed initialization of landmarks, and is interested in state estimation for robots with  particularly large dynamics and degrees of freedom, such as humanoids and aerial manipulators.
	His current projects turn around whole-body estimation and control, including multisensor fusion, localization and mapping, machine learning, and model predictive control.
\end{IEEEbiography}

\begin{IEEEbiography}[{\includegraphics[width=1in,height=1.25in,clip,keepaspectratio]{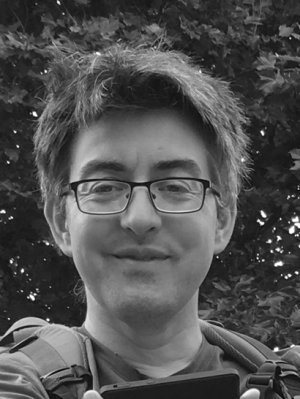}}]
	{Nicolas Mansard} received the M.Sc. degree in computer science from the University of Grenoble, Grenoble, France, in 2003 and the Ph.D. degree in robotics from the University of Rennes, Rennes, France, in 2006.

	He has been a CNRS Researcher since 2009.
	He was then Postdoctoral Researcher at Stanford University, Stanford, CA, USA with O. Khatib in 2007 and in JRL-Japan with A. Kheddar in 2008.
	He was Invited Researcher at the University of Washington with E. Todorov in 2014.
	He received the CNRS Bronze Medal in 2015 (one medal is awarded in France in automatic/robotic/signal-processing every year).
	His main research interests include the motion generation, planning and control of complex robots, with a special regard in humanoid robotics.
	His expertise covers sensor-based (vision and force) control, numerical mathematics for control, bipedal locomotion and locomotion planning.
	He published more than 70 papers in international journals and conferences and supervised 10 Ph.D. theses.

	Dr. Mansard is currently an Associate Editor of the \textsc{IEEE Transactions on Robotics}.
\end{IEEEbiography}

\begin{IEEEbiography}[{\includegraphics[width=1in,height=1.25in,clip,keepaspectratio]{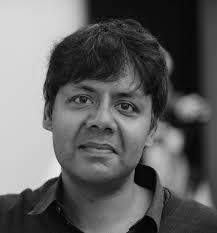}}]
	{Sethu Vijayakumar} received the Ph.D. degree in computer science and engineering from the Tokyo Institute of Technology, Tokyo, Japan, in 1998.

	He is Professor of Robotics and Founding Director of the Edinburgh Centre for Robotics, where he holds the Royal Academy of Engineering Microsoft Research Chair in Learning Robotics within the School of Informatics at the University of Edinburgh, U.K.
	He also has additional appointments as an Adjunct Faculty with the University of Southern California, Los Angeles, CA, USA and a Visiting Research Scientist with the RIKEN Brain Science Institute, Tokyo.
	His research interests include statistical machine learning, whole body motion planning and optimal control in robotics, optimization in autonomous systems as well as optimality in human motor motor control and prosthetics and exoskeletons.
	Professor Vijayakumar is a Fellow of the Royal Society of Edinburgh. In his recent role as the Programme Director for Artificial Intelligence and Robotics at The Alan Turing Institute, Sethu helps shape and drive the UK national agenda in Robotics and Autonomous Systems. 
\end{IEEEbiography}

\end{document}